\documentclass[11pt]{article}

\usepackage{appendix}

\usepackage[figuresright]{rotating}

\usepackage{tikz}
\usepackage{pgfplots}
\usepgfplotslibrary{groupplots} 
\usetikzlibrary{pgfplots.groupplots}

\usepackage{pgfplots,pgfplotstable}
\usepgfplotslibrary{fillbetween}
\pgfplotsset{compat=1.10}
\usepackage{amsmath,amssymb,ifthen,color,framed,bm}

\usepackage{graphicx}
\usepackage{dcolumn}
\usepackage{bm}
\usepackage{amscd,amsthm}
\usepackage{ifthen}
\usepackage{booktabs}

\usepackage{enumerate}

\usepackage{hyperref}
\hypersetup{
    bookmarks=true,         
    unicode=false,          
    pdftoolbar=true,        
    pdfmenubar=true,        
    pdffitwindow=false,     
    pdfstartview={FitH},    
    pdfsubject={Subject},   
    pdfproducer={Producer}, 
    pdfnewwindow=true,      
    colorlinks=true,       
    linkcolor=red,          
    citecolor=green,        
    filecolor=magenta,      
    urlcolor=cyan           
}

\newtheorem{lemma}{Lemma}

\newtheorem{theorem}{Theorem}

\newtheorem{algorithm1}{Algorithm}

\usetikzlibrary{matrix,arrows,decorations.pathmorphing}
\usepackage[vlined,ruled,linesnumbered]{algorithm2e}
\usepackage[
backend=bibtex,
url=false,isbn=false,doi=false,
date=year,
hyperref=auto,
style=alphabetic,
natbib=true,
sorting=nty,
firstinits=true,
maxnames=2, maxbibnames=10,
]{biblatex}

\bibliography{../bibliography.bib} 

\definecolor{DarkBlue}{rgb}{0,0,0.7}

\newcommand\Tinnerprod{}
\newcommand{\innerprod}[3][\Tinnerprod]{\ifthenelse{\equal{#1}{}}{\ensuremath{\left<#2,#3\right>}}{\ensuremath{\left<#2,#3\right>_{#1}}}}

\newcommand\Tex{}
\newcommand\PR[2][\Tex]{
\ifthenelse{\equal{#1}{}}{{\mathbb P}\left[#2\right]}{\ensuremath{{\mathbb P}_{#1}\left[ #2\right]}}}
\newcommand\EX[2][\Tex]{
\ifthenelse{\equal{#1}{}}{{\mathbb E}\left[#2\right]}{\ensuremath{{\mathbb E}_{#1}\left[ #2\right]}}}

\newcommand\defeq{:=}

\newcommand\vect[1]{\ensuremath{#1}}

\newcommand{\vw}{\vect{w}}

\newcommand{\ent}{H} 

\newcommand{\Pmat}{\ensuremath{M}}

\newcommand\setM{\mathcal M}

\newcommand{\reals}{\mathbb R}

\newcommand{\mc}{\mathcal}

\newcommand{\inv}[1]{  {#1}^{ -1 } } 

\newcommand{\h}{h}

\long\def\comment#1{}

\renewcommand{\S}{\mathcal S} 

\newcommand\score{\tau}
\newcommand\setb{k} 
\newcommand\numitems{n}
\newcommand{\Shat}{\ensuremath{\widehat{\S}}}
\newcommand{\setind}{\ensuremath{\ell}} 
\newcommand\numsets{\ensuremath{L}} 

\newcommand\alg{\mc A} 
\newcommand\stoptime{\xi} 
\newcommand\numcmp{N} 

\newcommand{\loindmone}[1]{{\setb_{#1-1}}}      
\newcommand{\upindpone}[1]{{\setb_{#1}+1}}	

\newcommand\gap[2]{\Delta_{#1,#2}}

\newcommand\gapup[2]{\bar{\Delta}_{#1,#2}}
\newcommand\gaplo[2]{{\underline{\Delta}}_{#1,#2}}

\newcommand\timeind{t}

\newcommand\bernrv{X} 
\newcommand\kl{d} 
\newcommand\KL{\mathrm{KL}} 

\newcommand\HD{\mathrm{D}} 
\newcommand\HammingD{\mathrm{D}_H} 
\newcommand\card[1]{\left|#1\right|}

\newcommand{\m}{m} 

\usepackage{dsfont}
\newcommand\ind[1]{\mathds{1}\{#1\}}
\newcommand{\event}{\ensuremath{\mathcal{E}}}

\newcommand{\scorehat}{\ensuremath{\widehat{\score}}}

\newcommand{\MYCDF}{\ensuremath{\Phi}}
\newcommand{\mypdf}{\ensuremath{\phi}}

\newcommand{\sigmaalgebra}{\mc F}

\newcommand{\comparisonclass}[1][]{
\ifthenelse{\equal{#1}{}}{\mc C}{\mc C_{#1}} }
\newcommand{\parfamily}[1][]{
\ifthenelse{\equal{#1}{}}{\mc C_{\mathrm{PAR}(\MYCDF)}}{\mc C_{\mathrm{PAR}(\MYCDF, #1)} }}

\newcommand\parw{w} 

\newcommand{\pmat}{\ensuremath{M}}

\newcommand{\pmatmin}{\pmat_{\mathrm{min}}}
\newcommand{\pdfmin}{\mypdf_{\mathrm{min}}}
\newcommand{\pdfmax}{\mypdf_{\mathrm{max}}}

\newcommand{\comparisonclasstil}{\comparisonclass} 

\newcommand\constup{c_{\mathrm{up}}}
\newcommand\clow{c_{\mathrm{low}}}

\newcommand{\FAR}{\ensuremath{f_{\tiny{\mbox{AR}}}}}
\newcommand{\FLOW}{\ensuremath{f_0}}
\newcommand{\FGEN}{\ensuremath{f}} 

\newcommand\eventscore{\event_\alpha} 

\newcommand{\SHORTSCORE}{\ensuremath{\{ \score_i\}_{i=1}^\numitems }}

\newcommand\complexityP{H}

\newcommand\mystackrel[2]{\stackrel{\text{#1}}{#2}}

\renewcommand\it{{a}} 
\newcommand\alt{b} 
\newcommand\alttwo{b'} 

\newcommand\NC{T}

\newcommand\NCb{\tilde \NC}

\newcommand\bup{b}
\newcommand\blow{b}

\newcommand\nextind{i} 
\newcommand\empgap{\widehat{\mathrm{gap}}}

\newcommand\constone{c_1}

\newcommand\dind{d}
\newcommand\mind{m}
\newcommand\bad{\event_{\text{bad}}}

\newcommand\Nup[2]{N^{\mathrm{up}}_{#1}(#2)}
\newcommand\Nlow[2]{N^{\mathrm{low}}_{#1}(#2)}
\newcommand\Ohidinglogs{\widetilde{O}}

\makeatletter
\long\def\@makecaption#1#2{
        \vskip 0.8ex
        \setbox\@tempboxa\hbox{\small {\bf #1:} #2}
        \parindent 1.5em  
        \dimen0=\hsize
        \advance\dimen0 by -3em
        \ifdim \wd\@tempboxa >\dimen0
                \hbox to \hsize{
                        \parindent 0em
                        \hfil 
                        \parbox{\dimen0}{\def\baselinestretch{0.96}\small
                                {\bf #1.} #2
                                } 
                        \hfil}
        \else \hbox to \hsize{\hfil \box\@tempboxa \hfil}
        \fi
        }
\makeatother

\begin{document}

\begin{center}

{\bf{\LARGE{
Approximate Ranking from Pairwise Comparisons
}}}

\vspace*{.2in}

{\large{
\begin{tabular}{cccc}
Reinhard Heckel$^{\ast}$ & Max Simchowitz$^{\star}$ & Kannan
Ramchandran$^{\star}$ & Martin J. Wainwright$^{\dagger,\star}$ \\
\end{tabular}
}}

\vspace*{.05in}

\begin{tabular}{c}
Department of ECE$^\ast$, Rice University
\end{tabular}

\begin{tabular}{c}
Department of Statistics$^\dagger$, and EECS$^\star$, 
University of California, Berkeley
\end{tabular}

\vspace*{.1in}

\today

\vspace*{.1in}


\begin{abstract}
A common problem in machine learning is to rank a set of $\numitems$
items based on pairwise comparisons.  Here ranking refers to
partitioning the items into sets of pre-specified sizes according to
their scores, which includes identification of the top-$\setb$ items
as the most prominent special case.  The score of a given item is
defined as the probability that it beats a randomly chosen other item.
Finding an exact ranking typically requires a prohibitively large
number of comparisons, but in practice, approximate rankings are often
adequate. Accordingly, we study the problem of finding approximate
rankings from pairwise comparisons.  We analyze an active ranking
algorithm that counts the number of comparisons won, and decides
whether to stop or which pair of items to compare next, based on
confidence intervals computed from the data collected in previous
steps.  We show that this algorithm succeeds in recovering approximate
rankings using a number of comparisons that is close to optimal up to
logarithmic factors.  We also present numerical results, showing that
in practice, approximation can drastically reduce the number of
comparisons required to estimate a ranking.
\end{abstract}

\end{center}

\section{Introduction}

The problem of ranking a collection of $\numitems$ items from noisy
pairwise comparisons arises in a wide range of applications, including
recommender systems for rating movies, books, or other consumer
items~\cite{piech_tuned_2013,aggarwal_recommender_2016}; peer grading
for ranking students in massive open online
courses~\cite{shah2013case}; ranking players in tournaments; search
engines; quantifying people's perception of cities from pairwise
comparison of street views of the
cities~\cite{salesses_collaborative_2013}; and online sequential
survey sampling for assessing the popularity of proposals in a
population of voters~\cite{salganik_wiki_2015}.

In each of these applications, the aim is to obtain a statistically
sound ranking from as few comparisons as possible.  In this work, we
investigate the power of adaptively selecting which pairs to compare
based on the outcomes of previous comparisons, a setting we call
\emph{active} or \emph{adaptive} ranking.  In contrast, \emph{passive}
or \emph{non-adaptive} ranking approaches fix the comparisons to make
before any data is collected.  It is well understood that one can
typically learn a ranking using fewer adaptively chosen comparisons
than one would need when passively choosing
comparisons~\cite{heckel_active_2016}.  However, for moderately large
or large collections of items--such as the ones that appear in most of
the applications mentioned above--or for collections with many items
of ``similar quality'' (to be made rigorous below), learning the
\emph{exact} ground-truth ranking may still require prohibitively many
comparisons.

Motivated by these large-scale ranking problems, this work studies the
problem of adaptivity obtaining \emph{approximate} rankings. We
demonstrate that learning an \emph{approximate} ranking may still be
statistically tractable even when recovering the \emph{exact} ranking
is not.  Formally, we consider a collection of $\numitems$ items, and
make comparison queries between pairs of items \mbox{$i, j \in
  [\numitems] \defeq \{1,2,\dots, \numitems\}$.} We assume that the
response to those queries are stochastic, where the probability that
item $i$ ``beats'' item $j$ is given by $\pmat_{ij} \in (0,1)$. We
assume that the outcomes of all queries are statistically independent,
and assume that either item $i$ or item $j$ ``wins'' the comparison
with probability $1$, which means that $\pmat_{ij} +\pmat_{ji} = 1$
for all $i \ne j$. Our aim is to rank the items in terms of their
Borda scores~\cite{de1781memoire}, defined as the probability that
item $i$ defeats an item chosen uniformly at random from $[n]
\setminus \{i\}$:
\begin{align}
\label{EqnDefnScore}
\score_i \defeq \frac{1}{\numitems - 1} \sum_{j\neq i} \pmat_{ij}~.
\end{align}
Apart from their intuitive appeal, the Borda scores generalize the
orderings considered in several popular comparison models, including
the classical, parametric Bradley-Terry-Luce
(BTL)~\cite{bradley_rank_1952,luce_individual_1959} and
Thurstone~\cite{thurstone_law_1927} models, as well as the
non-parametric Strong Stochastic Transitivity (SST)
model~\cite{tversky_substitutability_1969}. In all of these models,
the intrinsic model-defined ordering coincides with that given by the
scores $\SHORTSCORE$.  Rather than learning the scores $\SHORTSCORE$
exactly, or ranking items according to their exact score, this paper
considers the problem of \emph{approximately} partitioning the items
into sets of pre-specified sizes according to their respective scores.
This includes finding a total ordering that is approximately correct,
and the task of finding a set of $\setb$ items that is close to the
top-$\setb$ items.  For simplicity, we exclusively focus on the latter
problem in this paper.
 
\paragraph{Contributions:} Our main contribution is to present and analyze a novel active ranking algorithm for estimating an \emph{approximate} ranking of the items. 
The algorithm is based on adaptively estimating the scores to within
sufficient resolution to deduce a ranking.  We establish that with
high probability, the algorithm returns a ranking which satisfies the
desired approximation guarantee, and attains a distribution-dependent
sample complexity which can be parameterized in terms of the scores
$\SHORTSCORE$.  We then prove distribution-dependent lower bounds that
match our upper bound up to logarithmic factors for many problem
instances.  Our analysis leverages the fact that ranking in terms of
the scores $\SHORTSCORE$ is related to a particular class of
multi-armed bandit problems~\cite{even-dar_action_2006,
  bubeck_multiple_2013, urvoy_generic_2013}; this same connection has
been observed in the context of finding the top
item~\cite{yuekarmed2012, jamieson_sparse_2015,urvoy_generic_2013}.
Since to the best of our knowledge, the \emph{approximate} subset
selection problem has not been studied in the bandit literature, a
version of our algorithm and results are also new when specialized to
the multi-armed bandit problem.  Finally, we examine pathological
distributions for which the complexity of approximate ranking (or
approximate subset selection in the multi-armed bandit setup) seems to
diverge from what one would expect.  In these cases, we show that
careful randomized guessing strategies can yield significant
improvements in sample complexity.

\begin{figure}
\begin{center}
%
%
\includegraphics{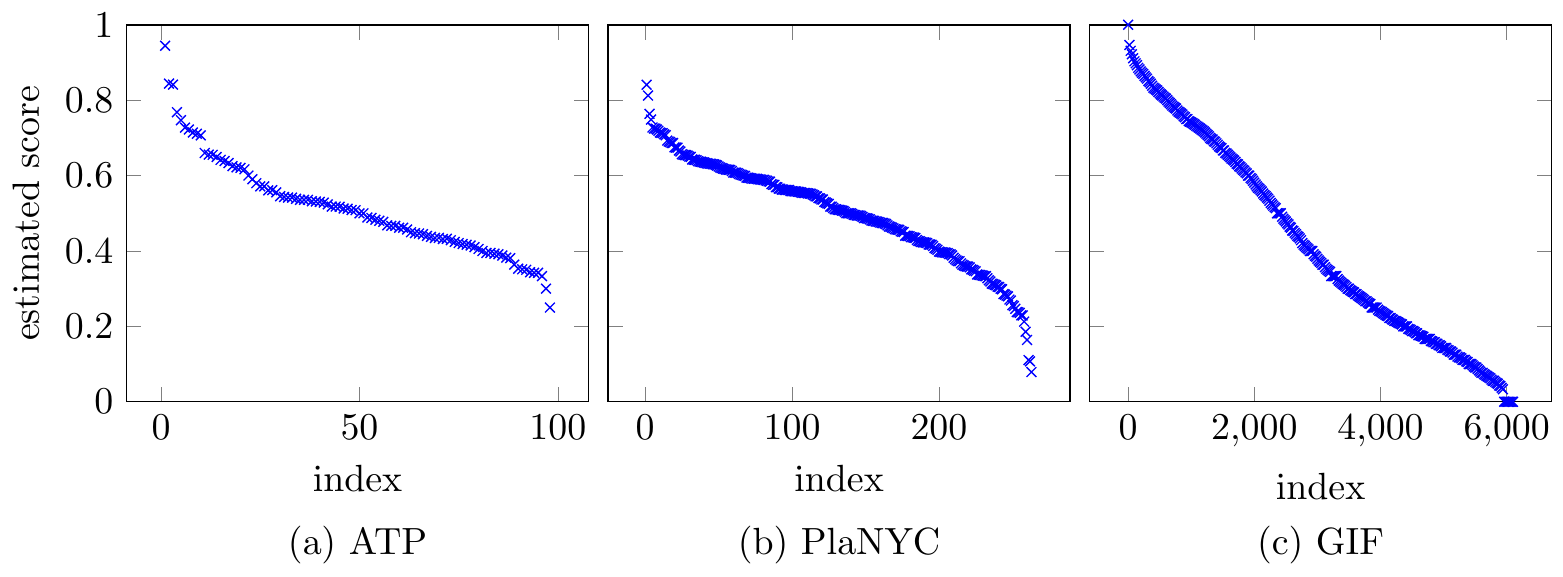}
\end{center}

\caption{\label{fig:borda}
Estimated scores from three different domains:
(a) Scores of the Association of Tennis Professionals (ATP) world tour, computed from the games played within a 52 week interval as the fraction of games won by the total number of games played. 
(b) Comparisons of the proposals in the PlaNYC survey, as reported in the paper~\cite{salganik_wiki_2015}
(only scores of items (proposals) that were rated at least $50$ times are
depicted). 
(c) Scores from comparisons of Gif's according to whether they display a certain emotion (see \href{http://www.gif.gf/}{http://www.gif.gf/}).
%
%
}
\end{figure}

\paragraph{Motivation for Approximate Rankings:}
In order to understand how approximation can drastically reduce the
number of comparisons required, let us consider a motivating example.
Suppose that we are interested in identifying the top-$\setb$ items,
and suppose for simplicity that the items are ordered, i.e., $\score_1
> \score_2 > \ldots > \score_\numitems$ (of course this ordering is
not known a-priori).  The paper~\cite{heckel_active_2016} shows that
in the active setting, the number of comparisons necessary and
sufficient for finding the top $k$ items is of the order
\begin{align}
\label{eq:active_exact} 
\sum\nolimits_{i=1}^{k} \frac{1}{ (\score_i - \score_{k+1})^{2}} 
+
\sum\nolimits_{i=k+1}^\numitems \frac{1}{ (\score_k - \score_i)^{2}},
\end{align}
up to a logarithmic factor. 
Thus, the sample complexity depends on the distribution of the scores;
see Figure~\ref{fig:borda} how these scores are distributed in some
applications. In practice, the differences between the scores often
obey the scaling $\score_i - \score_{i+1} \approx 1/\numitems$ on
average (see Figure~\ref{fig:borda}). To identify the top-$\setb$
items exactly, the aforementioned optimal active scheme would require
on the order of $\numitems^2$ comparisons, and a minimax-optimal
passive ranking scheme would even require on the order of
$\numitems^3$ comparisons~\cite{shah_simple_2015}.

Theorem~\ref{thm:LUCBHAMsuff} in this
paper shows that if one does not need to extract the exact top-$\setb$
items, but is instead willing to tolerate a few--say, $\h$
many--mistakes, then the number of comparisons shrinks drastically,
specifically by a factor proportional to $\h$.  In particular, if we
want to find a set $\S$ of $10\%$ of the items ($\setb =
0.1\numitems$) such that all but $10\%$ of the elements of $\S_1$ are
among the \emph{true} top $10\%$ of items ($\h = 0.1\setb$, $\setb =
0.1\numitems$), then the overall number of comparisons required would
be on the order of $\numitems^2/\h = 100\numitems$. Thus, relaxing to
approximate ranking can yield speedups that are \emph{linear} and
\emph{quadratic} in the number of items, compared to optimal exact
active and exact passive schemes. Moreover, our algorithm
(Algorithm~\ref{alg:HLUCB} below) that obtains this factor-of-$\h$
speedup does not require priori information about the spacings of the
$\SHORTSCORE$, but instead learns a near-optimal measurement
allocation for these scores adaptively.

\paragraph{Related works: }
There is a vast literature on ranking and estimation from pairwise
comparison data; however, most work focuses on finding \emph{exact}
rankings.  There are a number of papers~\cite{hunter2004mm,
  negahban_iterative_2012, hajek2014minimax, shah_estimation_2015,
  shah_simple_2015} devoted to settings in which pairs to be compared
are chosen a priori, whereas here we assume that the pairs may be
chosen in an active manner.  Moreover, several works impose
restrictions on the pairwise comparison probabilities, e.g., by
assuming the Bradley-Terry-Luce (BTL) parametric model (discussed
below) \cite{szorenyi_online_2015, hunter2004mm,
  negahban_iterative_2012, hajek2014minimax, shah_estimation_2015}.
\citet{eriksson_learning_2013} considers the problem of finding the
very top items using graph-based techniques, whereas
\citet{busa-fekete_top-k_2013} consider the problem of finding the
top-k items.  \citet{ailon_active_2011} considers the problem of
linearly ordering the items so as to disagree in as few pairwise
preference labels as possible.  Our work is also related to the
literature on multi-armed bandits, as discussed later in the paper.

\section{Problem formulation and background}

In this section we formally state the approximate ranking problem considered in this paper.

\subsection{Pairwise probabilities and scores}

Given a collection of items $[\numitems] \defeq
\{1,\ldots,\numitems\}$, let us denote by $\pmat_{ij} \in (0,1)$ the
(unknown) probability that item $i$ wins a comparison with item
$j$. We let $X_{ij}$ denote a Bernoulli random variable taking a value
of $1$ if $i$ beats $j$ and $0$ otherwise, so that $\pmat_{ij} =
\mathbb{E}[X_{ij}]$.  Moreover, we require that any comparison results
in a winner, so that $\pmat_{ij} + \pmat_{ji} = 1$.  For each item $i
\in [\numitems]$, recall that the score~\eqref{EqnDefnScore} defined
by $\score_i \defeq \frac{1}{\numitems-1} \sum_{j \in [\numitems]
  \backslash \{i\}} \pmat_{ij}$ corresponds to the probability that
item $i$ wins a comparison with an item $j$ chosen uniformly at random
from $[\numitems] \setminus \{i\}$.
We let $\pi \colon [\numitems]\rightarrow [\numitems]$ denote any
(possibly non-unique) permutation such that $ \score_{\pi(1)} \ge
\score_{\pi(2)} \ge \ldots \ge \score_{\pi(\numitems)}.  $ In words,
$\pi(i)$ denotes the item with the $i^{th}$ largest score.  Ranking
corresponds to partitioning the items into disjoint sets according to
its scores.  For simplicity, in this paper we focus on the ranking
problem of splitting $[\numitems]$ into the top-$\setb$ items and its
complement \mbox{$\S_1 \defeq \{\pi(1), \ldots, \pi(\setb)\},\;\; \S_2
  \defeq \{\pi(\setb+1), \ldots, \pi(\numitems) \}$.} In this work,
our goal is to find an approximation to $\S_1$ and $\S_2$ in terms of
the Hamming distance between two sets $\S,\S'$, defined as
\mbox{$\HammingD(\S,\S') \defeq \card{ ( \S \cup \S')\setminus (\S
    \cap \S') }$.}  Specifically, we say the ranking $\Shat_1,\Shat_2$
with $|\Shat_\setind| = |\S_\setind|$ is $\h$-Hamming-accurate if
\begin{align*}
\HammingD(\Shat_\setind,\S_\setind) \leq 2\h, \quad \text{ for } \ell
\in \{1,2\}.
\end{align*}
For future reference, we define
\begin{align*}
\comparisonclass[\pmatmin] \!\defeq \!\left\{\Pmat \in (0,1)^{n\times
  n} \!\mid\! \pmat_{ij} \!=\! 1\!-\!\pmat_{ji}, \pmat_{ij} \!\geq\!
\pmatmin \right\},
\end{align*}
corresponding to the set of pairwise comparison matrices with pairwise
comparison probabilities lower bounded by $\pmatmin$.


\subsection{The active approximate ranking problem}

An active ranking algorithm acts on a pairwise comparison model $\pmat
\in \comparisonclass[0]$.  The goal is to obtain an approximate
partition of the items into disjoint sets from active comparisons.  At
each time instant, the algorithm can compare two arbitrary items,
which the algorithm may select based on the outcomes of previous
comparisons.  When comparing $i$ and $j$, the algorithm obtains an
independent draw of the random variable $\pmat_{ij}$ in response.  The
algorithm terminates based on an associated stopping rule, and returns
an approximate ranking $\Shat_1,\Shat_2$.  For a given tolerance
parameter $\delta \in (0,1)$, we say a ranking algorithm $\alg$ is
$(\h,\delta)$-accurate for a pairwise comparison matrix $\pmat$, if
the ranking returned is $\h$-Hamming accurate with probability at
least $1-\delta$.  Moreover, we say that $\alg$ is \emph{uniformly
  $(\h,\delta)$-accurate} over a given set of pairwise comparison
models $\comparisonclasstil$ if it is $\delta$-accurate for each
$\Pmat \in \comparisonclasstil$.


\subsection{Relation to multi-armed bandits}

The \emph{exact} version of the ranking problem considered in this
paper is related to the subset selection problem in the bandit
literature~\cite{kalyanakrishnan_pac_2012}.  Specifically, a
multi-armed bandit model consists of $\numitems$ arms, each a random
variable with unknown distribution.  The subset selection problem is
concerned with identifying the top arms (according to the means) by
taking independent draws of the random variables.  Various works
\cite{yue_beat_2011,yuekarmed2012,urvoy_generic_2013,jamieson_sparse_2015}
have observed that, by definition of the score $\score_i$, comparing
item $i$ to an item chosen uniformly at random from
$[\numitems]\setminus \{i\}$ can be modeled as drawing a Bernoulli
random variable with mean $\score_i$.  Our subsequent analysis relies
on this relation.

However, when viewing our problem as a multi-armed bandit problem with
means $\{\score_i\}_{i=1}^\numitems$, we are ignoring the fact that
the means are coupled, as they must be realized by some pairwise
comparison matrix $\pmat$.  Due to $\pmat_{ij} = 1 - \pmat_{ji}$, this
matrix must satisfy certain constraints, such as $\sum_{i=1}^\numitems
\score_{i} = \numitems/2$ and $\sum_{i=1}^j \score_{\pi(i)} \geq
\frac{1}{\numitems - 1} \frac{j (j-1)}{2}$ (e.g., see the
papers~\cite{landau_dominance_1953,joe_majorization_1988}).  Our
algorithm turns out to be near-optimal, even though it does not take
those constraints into account. This seems to corroborate the
observation in~\cite{simchowitz_simulator_2017} that many types of
constraints surprisingly do not improve the sample complexity of
bandit problems.

Finally, at least to the best of our knowledge, the problem of
\emph{approximate} subset selection has not been studied in the bandit
literature, meaning that our algorithm and results are also new when
specialized to the multi-armed bandit problem.  However, it should be
noted that other versions of approximation have been considered in the
literature; for instance, \citet{zhou_optimal_2014} studied the
problem of selecting $k$ arms with low aggregate regret, defined as
the gap between the average reward of the optimal solution and the
solution given by the algorithm.


\subsection{Parametric models}
\label{sec:intro_parametric}

In this section, we introduce a family of parametric models that are
popular in the pairwise comparison
literature~\cite{szorenyi_online_2015, hunter2004mm,
  negahban_iterative_2012, hajek2014minimax, shah_estimation_2015}. We
focus on these parametric models in
Section~\ref{sec:parmodlowerbounds}, where we show that, perhaps
surprisingly, if the pairwise comparison probabilities are bounded
away from zero, for most constellations of scores, these assumptions
can at most provide little gains in sample complexity.

Any member of this family is defined by a strictly increasing and
continuous function $\MYCDF \colon \reals \to [0,1]$ obeying $\MYCDF(
t ) = 1 - \MYCDF(-t)$, for all $t \in \reals$. The function $\MYCDF$
is assumed to be known.  A pairwise comparison matrix in this family
is associated to an unknown vector $\vw \in \reals^\numitems$, where
each entry of $\vw$ represents some quality or strength of the
corresponding item.  The parametric model $\parfamily$ associated with
the function $\MYCDF$ is defined as:
\begin{align*}
\parfamily = \{ \pmat_{ij} = \MYCDF(\parw_i - \parw_j) \ \ \forall i,j \in [\numitems],
  \vw \in
\reals^\numitems \}.
\end{align*}
Popular examples of models in this family are the Bradley-Terry-Luce
(BTL) model, obtained by setting $\MYCDF$ equal to the sigmoid
function $\MYCDF(\timeind) = \frac{1}{1 + e^{-t}}$, and the
Thurstone model, obtained by setting $\MYCDF$ equal to the Gaussian
CDF.  Since \mbox{$\score_1 > \score_2 > \ldots >
  \score_\numitems$} is equivalent to \mbox{$\parw_1 > \parw_2 >
  \ldots > \parw_\numitems$}, the ranking induced by the
 scores $\SHORTSCORE$ is equivalent to that induced by $\parw$.

\section{Hamming-LUCB: Algorithm and analysis}
\label{sec:hamming}

In this section, we present our approximate ranking algorithm, and an analysis proving that it is near optimal for many interesting and natural problem instances.

\subsection{The Hamming-LUCB algorithm}

Our algorithm is based on actively identifying sets $\tilde \S_1$ and $\tilde \S_2$ consisting of $\setb - \h$ items and $\numitems - \setb - \h$ items, respectively, such that with high confidence the items in the first set have a larger score than the items in the second set.
Once we have found such sets, we can arbitrarily distribute the remaining items to the sets $\tilde \S_1$ and $\tilde \S_2$ in order to obtain a Hamming-accurate ranking with high confidence. 

Our algorithm identifies those sets based on adaptively estimating the
scores $\SHORTSCORE$.  We estimate the score of item $i$ by comparing
item $i$ with items chosen uniformly at random from
$[\numitems]\setminus \{i\}$, which yields an unbiased estimate of
$\score_i$.  The key idea is to only estimate the scores sufficiently
well so we can obtain the two sets $\tilde \S_1$ and $\tilde \S_2$
from them.  This strategy decides based on the current estimates of
the scores and associated confidence intervals which estimate to
``update'', by comparing it to a randomly chosen item.  Our strategy
to update the estimates of the scores is guided by the insight that
the ``easiest'' items to distinguish are the top $\setb-\h$ items,
$\{\pi(1),\dots,\pi(\setb-\h)\}$, and the bottom $\numitems - \setb -
\h$ items, $\{\pi(\setb+\h+1),\dots,\pi(n)\}$.  Hence, our algorithm
focuses on what it ``thinks'' are those top and bottom items.

\begin{figure}
\begin{center}
\includegraphics{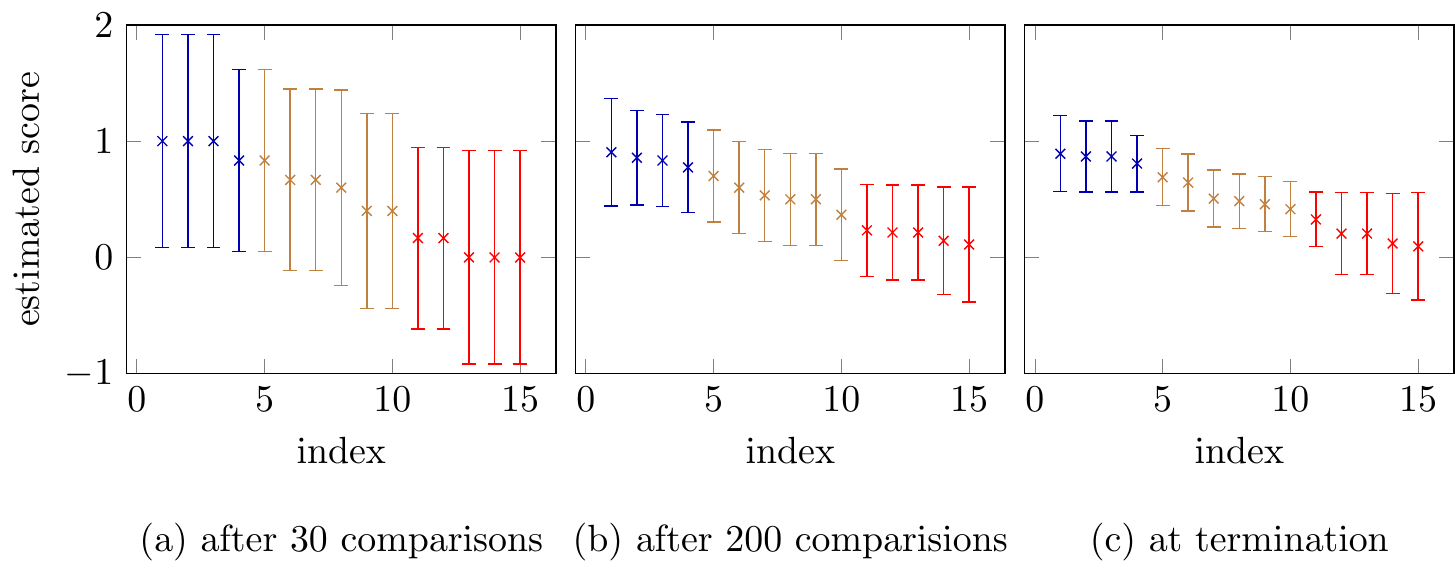}
\end{center}
\caption{\label{fig:algvis}
Visualization of a run of the Hamming-LUCB algorithm on a problem instance with scores evenly spaced in the interval $[0.1,0.9]$, and parameters $\setb=7,\h=3$. 
The estimates of the scores $\scorehat_i$ of the top items $\{(1), \ldots, (\setb-\h) \}$, the middle items $\{\pi(\setb-\h+1), \ldots, \pi(\setb+\h)\}$, and the bottom items $\{(\setb+1+\h), \ldots, (\numitems) \}$, along with the confidence intervals $[\scorehat_i - \alpha_i, \scorehat_i + \alpha_i]$ are depicted in blue, brown, and red, respectively, after 30 and 200 comparisons, and at termination.
Note that once the confidence intervals of the top and bottom items are separated, the algorithm terminates. 
}
\end{figure}

We define a confidence bound based on an non-asymptotic version of the
law of the iterated
algorithm~\cite{kaufmann_complexity_2014,jamieson_lil_2014}; it is of
the form \mbox{$\alpha(u) \propto \sqrt{ \frac{ \log( \log(u)
      \numitems/\delta) }{ u } }$,} where $u$ is an integer
corresponding to the number of comparisons, and with the constants
involved explicitly chosen by setting 
\[
 \alpha(u) = \sqrt{ \frac{
    \beta(u,\delta/\numitems) }{ 2u} }, \quad \text{with }
\beta(u,\delta') = \log(1/\delta') + 0.75\log\log(1/\delta') +
1.5\log(1+\log(u/2)).
\]
For each item $i \in [\numitems]$, the
algorithm stores a counter $\NC_i$ of the number of comparisons in
which it has been involved, along with an empirical estimate of the
associated score $\scorehat_i(\NC_i)$.  For notational convenience, we
adopt the shorthands $\scorehat_i = \scorehat_i(\NC_i)$ and $\alpha_i
= \alpha(\NC_i)$.  Within each round, we also let $(\cdot)$ denote a
permutation of $[\numitems]$ such that \mbox{$\scorehat_{(1)} \ge
  \scorehat_{(2)} \geq \dots \geq \scorehat_{(\numitems)}$.} We then
define the indices
\begin{align}
\label{eq:d_ind}
\dind_1 = \underset{i \in \{ (1), \ldots, (\setb - \h) \}}{\arg \min} \scorehat_i - \alpha_i,
\quad 
\dind_2 = \underset{i \in \{ (\setb+1+\h), \ldots, (\numitems) \}}{\arg \max} \scorehat_i + \alpha_i. 
\end{align}
These indices are the analogues of the standard indices of the Lower-Upper Confidence Bound (LUCB) strategy from the bandit literature \cite{kalyanakrishnan_pac_2012} for the top $\setb - \h$ and bottom $\numitems - \h - \setb$ items. 
The LUCB strategy for exact top $\setb$ recovery would update the scores $\dind_1$ and $\dind_2$ (for $\h=0$) at each round. 
As mentioned before, our strategy will go after what it ``thinks'' are the top $\setb-\h$ items, $\tilde \S_1 = \{(1), \ldots, (\setb-\h) \}$, 
and what it ``thinks'' are the bottom $\numitems-\setb-\h$ items, $\tilde \S_2 = \{(\setb+1+\h), \ldots, (\numitems) \}$.
Moreover, the algorithm keeps all the other items in consideration for inclusion in these sets, by keeping their confidence intervals below the confidence intervals of the items in $\tilde \S_1$ and $\tilde \S_2$ (cf.~equation~\eqref{eq:blowbupdef} in the algorithm below). 
This is crucial to ensure that the algorithm does not get stuck trying to distinguish the middle items $\{\pi(\setb-\h+1), \ldots, \pi(\setb+\h)\}$, which in general requires many comparisons, as their scores are typically closer. 
In Figure~\ref{fig:algvis} we show an example run of the Hamming-LUCB algorithm, to illustrate the idea. 

\begin{algorithm}[]
\textbf{Input:} Confidence parameter $\delta$.\\ 
\textbf{Initialization}: For every item $i \in [\numitems]$, compare $i$ to an item $j$ chosen uniformly at random from $[\numitems] \setminus \{i\}$, and set $\scorehat_i(1)=\ind{i \text{ wins}}$, $\NC_{i} = 1$. \\
\textbf{Do} until termination:\\
\Indp Let $(\cdot)$ denote a permutation of $[\numitems]$ such that $\scorehat_{(1)} \ge \scorehat_{(2)} \ge \dots \scorehat_{(\numitems)}$. \\
For $\dind_1$ and $\dind_2$ defined by equation~\eqref{eq:d_ind}, define the indices
\begin{align}
\label{eq:blowbupdef}
\blow_1 = \underset{ i \in \{ \dind_1,(k-h+1),\dots,(k)\} }{\arg \max} ~ \alpha_i,
\quad
\text{and}
\quad 
\bup_2 = \underset{i \in \{\dind_2,(k+1),\dots,(k+h)\}}{\arg \max} ~ \alpha_i. 
\end{align}
\\
\textbf{For} $\nextind \in \{\blow_1, \bup_2\}$, increment $\NC_\nextind\leftarrow \NC_{\nextind}+1$, 
compare $i$ to an item $j$ chosen uniformity at random from $[\numitems] \setminus \{i\}$, 
and update 
$\scorehat_{\nextind} \leftarrow \frac{\NC_\nextind-1}{\NC_\nextind} \scorehat_\nextind + 
\frac{1}{\NC_\nextind} \ind{ \nextind \text{ wins}}$. \\
 \textbf{End Loop} once the termination condition holds:
\begin{align}
\label{eq:termination}
\scorehat_{d_1} - \alpha_{d_1}
\geq 
\scorehat_{d_2} + \alpha_{d_2}.
\end{align}\\
\Indm
\textbf{Return} the estimates of the partitions 
$\Shat_1 = \{(1), \ldots, (\setb) \}$ and 
$\Shat_2 = \{(\setb+1), \ldots, (\numitems)\}$.

\caption{
\label{alg:HLUCB}
Hamming-LUCB}
\end{algorithm}

\subsection{Guarantees and optimality of the Hamming-LUCB algorithm}

We next establish guarantees on the number of comparisons for the Hamming-LUCB algorithm to succeed. 
As we show below, the number of comparisons depends on the following gaps between the scores 
\begin{align*}
\gap{i}{\setb+1+\h}
\quad
\text{and}
\quad
\gap{\setb-\h}{i},
\quad 
\text{ where }
\quad
\gap{i}{j} \defeq \score_i - \score_j.
\end{align*}
Thus, as one might intuitively expect, the number of comparisons is typically smaller when $\h$ is larger, as the corresponding gaps typically become larger.

\begin{theorem}
\label{thm:LUCBHAMsuff}

For any $\Pmat \in \comparisonclass[0]$, the Hamming-LUCB algorithm run with confidence parameter $\delta$ is $(\h,\delta)$-Hamming-accurate,  and with probability at least $1-\delta$, makes at most $\Nup{\h}{\pmat}$ comparisons, where
\begin{align}
\label{eq:sampcsimple}
\Nup{\h}{\pmat}
=
\Ohidinglogs\left(
\sum_{i = 1}^{\setb-\h}
\gap{i}{\setb+1+\h}^{-2}
+
\sum_{i = \setb+1+\h}^{\numitems} 
\gap{\setb-\h}{i}^{-2}
+
2\h \gap{\setb-\h}{\setb+1+\h}^{-2}
\right).
\end{align}
The notation $\Ohidinglogs$ absorbs factors logarithmic in $\numitems$, and doubly logarithmic in the gaps. 
\end{theorem}

Theorem~\ref{thm:LUCBHAMsuff} proves that the Hamming-LUCB algorithm is $(\h,\delta)$-accurate, and characterizes the number of comparisons that it requires as a  function of the gaps between the scores.

Comparing $\Nup{\h}{\pmat}$ to the number of comparisons necessary and sufficient for finding the top-$\setb$ items, we see that the Hamming-LUCB algorithm depends on the gaps $\gap{i}{\setb+1+\h}$ and $\gap{\setb-\h}{i}$ instead of the gaps $\gap{i}{\setb+1}$ and $\gap{\setb}{i}$ which appear in the sample complexity for finding the top $\setb$ items (cf. equation~\eqref{eq:active_exact}). 
These gaps are typically significantly larger, resulting in a lower sample complexity. 
For example, in practice, the scores are often increasing in that $\score_i-\score_{i+1}$ is on average on the order of $1/\numitems$. 
Thus, for sufficiently large $\h$, several real world models belong to the class (see Figure~\ref{fig:borda} for plausible members of this class):
\begin{align}
\comparisonclass[\beta,\h]
\defeq 
\left\{\Pmat \in (0,1)^{n\times n}
\mid \pmat_{ij} = 1-\pmat_{ji}, 
\text{ and } \tau_i - \tau_{i+\h} \geq \beta \h / \numitems, \text{ for all } i \right\}.
\label{eq:defcomparisonclassalpha}  
\end{align}
For this class, the complexity of finding the top-$\setb$ items with the Hamming-LUCB algorithm is on the order of $\Ohidinglogs\left( \numitems^2/ (\beta^2 \h )\right)$, which is by a factor of $\h$ smaller than the complexity for finding the exact top-$\setb$ items. 

Moreover, Hamming LUCB provides a strict improvement over the optimal sample complexity in the passive setup, for which Shah and Wainwright~\cite{shah_simple_2015} establish upper bounds and minimax lower bounds which state that  $O(\numitems \log \numitems / \gap{\setb-\h}{\setb+1+\h}^2)$ comparisons 
are necessary and sufficient to identify the top $\setb$ items up to a Hamming error $\h$ with high probability. 

As $\h$ increases, the upper bound depends on gaps between items with increasingly disparate position in the ranking, and thus, the upper bound on the sample complexity decreases. The following lower bound shows that, up to logarithmic factors in $\numitems$, doubly logarithmic factors in the gaps, and a multiplicative scaling of $\h$, the Hamming-LUCB algorithm is optimal. 

\begin{theorem}\label{thm:lowerbound}
For any $\delta \in (0, 0.14]$, let $\alg$ denote an algorithm which is uniformly $(\h,\delta)$-accurate over $\comparisonclass[1/8]$. 
Then, when $\alg$ is run on any comparison instance $\Pmat \in
  \comparisonclass[3/8]$, $\alg$ must make at least $\Nlow{\h}{\pmat}$ comparisons in expectation, where 
\begin{align*}
\Nlow{\h}{\pmat} \defeq \clow \log\left(\frac{1}{2 \delta}
\right) 
\left(
\sum_{i = 1}^{\setb - 2\h} \gap{i}{\setb+1+2\h}^{-2}
+ 
\sum_{i = \setb+1+2\h}^{\numitems} \gap{\setb - 2\h}{ i }^{-2}
\right),
\end{align*}
for some universal constant $\clow$. 
\end{theorem}

Note that the above lower bound \emph{does not} depend on the gaps
involving the items $\setb - 2\h+1,\dots,\setb+2\h$. However, we can 
still relate the lower bound to the upper bound by (see Section~\ref{sec:proofequplorel} for the simple proof)
\begin{align}
\label{eq:UpperLowerBoundRel}
\Nup{3\h}{\pmat} 
\le 
\Ohidinglogs( \Nlow{\h}{\pmat} ),
\end{align}
so that we see that, up to rescaling our Hamming error tolerance $\h$, 
our upper and lower bounds ($\Nup{\h}{\pmat}$ and $\Nlow{\h}{\pmat}$, respectively) match up to logarithmic factors. 
For many problem instances of interest---such as models in the class $\comparisonclass[\beta,\h]$ in equation~\eqref{eq:defcomparisonclassalpha}---the sample complexity bounds 
$\Nup{3\h}{\pmat}$ and $\Nlow{\h}{\pmat}$ degrade gracefully with the Hamming tolerance $\h$, so that typically we have 
$\Nup{\h}{\pmat} \le \Ohidinglogs (\Nlow{\h}{\pmat})$.


Observe that if $\h = 0$, we recover the exact top-$\setb$ recovery upper bound in equation~\eqref{eq:active_exact}, 
which is related to similar results for multi armed bandits~\cite{kalyanakrishnan_pac_2012}. 
We believe that by modifying the confidence intervals in Hamming LUCB as in the LUCB++ algorithm of~Simchowitz et al.~\cite{simchowitz_simulator_2017}, one can sharpen the upper bound $\Nup{\h}{\pmat}$ on the sample complexity by replacing $\log \numitems$ with $\log \setb$ on the terms $\Delta_{\setb-\h,i}^{-2}$ corresponding to items $i \in \{\setb+\h+1,\dots,\numitems\}$, thereby matching known lower bounds for top-$\setb$ subset selection problem in the bandit literature~\cite{simchowitz_simulator_2017,chen2017nearly,kalyanakrishnan_pac_2012}. In the interest of simplicity, we defer refining these logarithmic factors to later work.

\subsection{Parametric models}

\label{sec:parmodlowerbounds}

Even though the lower bound of $\Nlow{\h}{\pmat}$ qualitatively matches the upper bound $\Nup{\h}{\pmat}$, it gives the misleading impression that an $\h$-approximate algorithm can get away without querying the items in $\{\setb-2\h,\dots,\setb+2\h+1\}$. In the proof section, we use techniques from~\cite{simchowitz_simulator_2017} and~\cite{chen2017nearly} to establish a more refined 
technical lower bound showing that all items, 
including those with ranks close to $\setb$ must be compared an ``adequate'' number of times. 
For simplicity, we state a consequence of this lower bound applied to the parametric models described in Section~\ref{sec:intro_parametric}.
In addition to showing that each item has to be compared a certain number of times, this bound also establishes that \emph{even knowledge of the exact parametric form of the pairwise comparison probabilities $\pmat$} cannot drastically improve the performance of an active ranking algorithm. 

In more detail, we say that a model is parametric, if there exists a strictly increasing CDF $\MYCDF\colon \reals \to [0,1]$ such that $\pmat_{ij} = \MYCDF(\parw_i - \parw_j)$ for some weights $\{\parw_j\}$. 
For any pair of constants $0 <
\pdfmin \leq \pdfmax < \infty$, we say that a CDF $\MYCDF$ is
$(\pdfmin,\pdfmax,\pmatmin)$-bounded, if it is differentiable, and if
its derivative $\MYCDF'$ satisfies the bounds
\begin{align}
\label{eq:assumptionderivativeCDF}
\pdfmin \leq \MYCDF'(\timeind) \leq \pdfmax, \quad \text{for all } t
\in [\inv{\MYCDF}(\pmatmin), \inv{\MYCDF}(1-\pmatmin)].
\end{align}
Note that for the popular BTL and Thurstone models, equation~\eqref{eq:assumptionderivativeCDF} holds with $\pdfmin/\pdfmax$ close to one, provided that $\pmatmin$ is not too small. 
We say that an algorithm is symmetric if its distribution of comparisons commutes with permutations of the items. For any such algorithm, our main lower bound is as follows:

\begin{theorem} 
\label{thm:parametric}
For a given $\delta \leq \frac{1}{2} \min(\frac{1}{\setb}, \frac{1}{\numitems-\setb})$, let $\alg$ be any symmetric algorithm 
that is uniformly $(\h,\delta)$-Hamming accurate over $\Pmat \in \parfamily \cap \comparisonclass[\pmatmin]$. 
Then, when $\alg$ is run on the instance $\Pmat \in \parfamily \cap \comparisonclass[\pmatmin]$, for any integer $q\geq 1$ and any item $\it \in [\numitems]$, it must make at least 
\begin{align*}
\frac{\pmatmin\pdfmin^2}{3 \pdfmax^2} 
\left( \frac{2q -1}{2\h + q} \right)^2
\max_{\alt \in 
\{\setb - 2(\h+q),\setb + 1 +  2(\h + q)\} 
}
\Delta_{a,\alt}^{-2}
\end{align*}
comparisons involving item $\it$ on average. 
\end{theorem}

In particular, by choosing $q=\h$, we see that the total sample complexity is lower bounded by
\begin{align}
\label{eq:totalparam}
\sum\nolimits_{i = 1}^{\setb-3\h}
\gap{i}{\setb+1+3\h}^{-2}
+
\sum\nolimits_{i = \setb+1+3\h}^{\numitems} 
\gap{\setb-3\h}{i}^{-2}
+
6\h \gap{\setb-3\h}{\setb+1+3\h}^{-2},
\end{align}
which is equivalent to the upper bound $\Nup{3\h}{\pmat}$ achieved by the Hamming-LUCB algorithm up to logarithmic factors. 
The lower bound from Theorem~\ref{thm:parametric} is stronger than the lower bound from Theorem~\ref{thm:lowerbound}, in that it 
applies to the larger class of algorithms that are only $(\h,\delta)$-accurate over the smaller class of parametric models. 
In fact, the parametric subclass $\parfamily \cap
\comparisonclass[\pmatmin]$ is significantly smaller than the full set of pairwise comparison models $\comparisonclass[\pmatmin]$, in the sense that one can find matrices in $\comparisonclass[\pmatmin]$ that cannot be
well-approximated by any parametric model~\cite{shah_stochastically_2015}. 
Therefore, theorem~\ref{thm:parametric} shows that, up to rescaling the Hamming error tolerance $\h$ and logarithmic factors, the Hamming-LUCB algorithm is optimal, even if we restrict ourself to algorithms that are uniformly $(\h,\delta)$-accurate \emph{only} over a parametric subclass.
Thus, in the regime where the pairwise comparison probabilities are bounded away from zero, parametric assumptions cannot substantially reduce the sample complexity of finding an \emph{approximate} ranking; an observation that has been made previously in the paper~\cite{heckel_active_2016} for exact rankings.
 
A second and equally important consequence of Theorem~\ref{thm:parametric} is that each item has to be sampled a certain number of times, an intuition not captured by Theorem~\ref{thm:lowerbound}. 
This conclusion continues to hold for general pairwise comparison matrices, please see Theorem~\ref{thm:lowerboundmoderateconf} in Section~\ref{sec:altcmp} for a formal statement. 

%
%
%


\subsection{Random guessing}

Even though our the upper and lower bounds essentially match whenever $\Nup{3\h}{\pmat} \approx \Nup{\h}{\pmat}$, there are there are pathological instances where $\Nup{3\h}{\pmat} \ll \Nup{\h}{\pmat}$, and where the Hamming-LUCB algorithm will make considerably more comparisons than a careful random guessing strategy.
 

As an example, consider a problem instance parameterized by $\kappa, \epsilon$, with scores given by 
\[
\score_i
=
\begin{cases}
1/2 + \kappa, & i \in \{1,\ldots, \setb - \h -2\} \\
1/2 + 2\epsilon , & i = \setb - \h - 1 \\
1/2 + \epsilon, & i \in \{\setb - \h, \ldots, \setb \} \\
1/2 - \epsilon, & i \in \{\setb +1, \ldots, \setb+1+\h \} \\
1/2 - 2\epsilon, & i = \setb+2+\h \\
1/2 - \kappa, & i \in \{\setb + \h + 2,\ldots, \numitems\} 
\end{cases},
\]
for some $\kappa$ and $\epsilon$. 
The upper bound~\eqref{eq:sampcsimple} for the Hamming-LUCB strategy is at least on the order of $\h / \epsilon^2$, since the gap between the $(\setb - \h)$-th and the $(\setb + 1 + \h)$-th largest score is $4\epsilon$. 
However, the lower bound provided by Theorem~\ref{thm:lowerbound} is 
$(\numitems - 2(\h+2))/\kappa^2$, which is independent of $\epsilon$. 
Thus, by making $\epsilon$ small, the ratio of upper and lower bounds becomes arbitrarily large. Intuitively, Hamming-LUCB is wasteful because it is attempting to identify the \emph{exact} top $\setb - \h$ arms with too much precision. However, for this particular problem instance, the following random guessing strategy will attain our lower bound. First, we obtain estimates $\scorehat_i$ of each score $\score_i$ by comparing item $i$ to 
$\NC_i = c \log (n/\delta) / \kappa^2$ randomly chosen items. For each score, test whether there are $\setb - \h - 2$ items obeying $\scorehat_i \geq 1/2 + c\sqrt{\log (\numitems) / \NC_i}$ and whether there are $\numitems - (\setb + \h + 2)$ items obeying
$\scorehat_i \leq 1/2 - c\sqrt{\log (\numitems) / \NC_i}$.
If yes, assign these items the estimates $\Shat_1$ and $\Shat_2$, respectively, and assign all remaining items uniformly at random to the sets $\Shat_1$ and $\Shat_2$, and terminate.



\section{Experimental results}

\begin{figure}
\begin{center}
%
%
\includegraphics{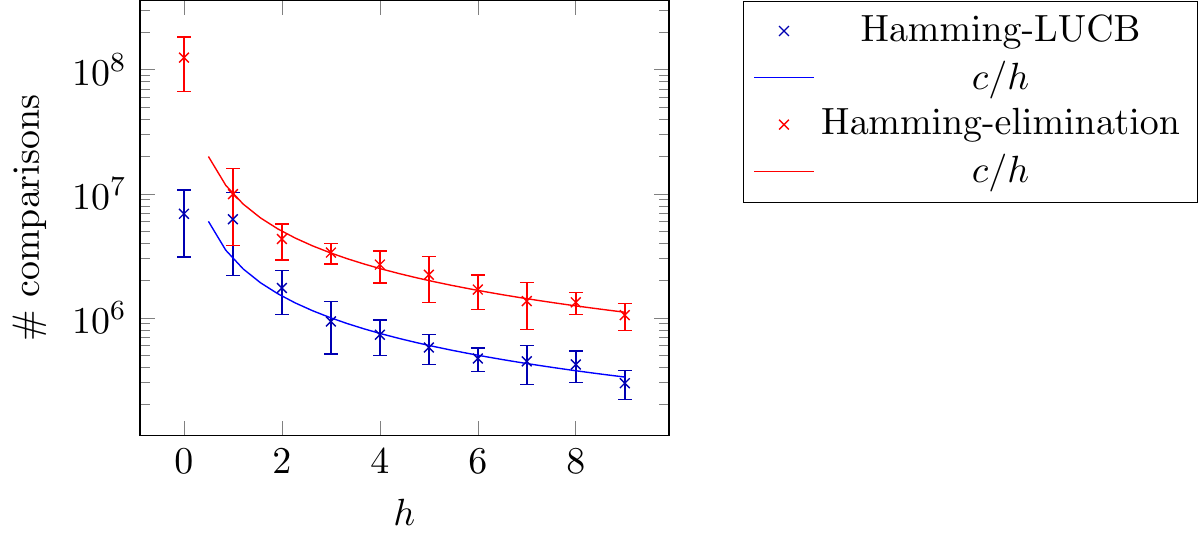}
\end{center}

\caption{\label{fig:cmp}
Sample complexity of the Hamming-LUCB algorithm and an elimination strategy run on a pairwise comparison model resembling the PlaNYC online sequential survey.
Both algorithms find the top $35$ proposals out of $263$ proposals, up to Hamming error $\h$.
The error bars correspond to one standard deviation from the mean. 
The results show that the sample complexity of the Hamming-LUCB algorithm for finding an $\h$-accurate ranking drops by a factor of about $\h$.
Moreover, the Hamming-LUCB algorithm requires significantly fewer samples than the elimination strategy.  
}
\end{figure}

In this section, we provide experimental evidence that corroborates our theoretical claims that the Hamming-LUCB algorithm allows to significantly reduce the number of comparisons if one is content with an approximate ranking. We show that these gains are attained on a real-world data set. 
Specifically, we generate a pairwise comparison model by choosing $\pmat$ such that the Borda scores $\score_i$ coincide with those found empirically in the PlaNYC survey~\cite{salganik_wiki_2015}; see panel (b) of Figure~\ref{fig:borda}. 
We emphasize that, since Hamming LUCB depends only on the Borda scores $\score_i$ and not on the comparison probabilities $\pmat_{ij}$, these simulations provide a faithful representation of how Hamming LUCB performs on real-world data.
%
%
In Figure~\ref{fig:cmp}, we plot the results of running the Hamming-LUCB algorithm on the PlanNYC-pairwise comparison model in order to determine the top $\setb=35$ items, for different values of $\h$.
We observed that the results for other values of $\setb$ are very similar. 
As suggested by our theory, the number of comparisons to find an approximate ranking decays in a manner inversely proportional in $\h$. 
%
%
We compare the Hamming-LUCB algorithm to another sensible active ranking strategy for obtaining an Hamming-accurate ranking. 
Specifically, we consider a version of the successive elimination strategy proposed in~\cite[Sec.~3.1]{heckel_active_2016} for finding an exact ranking. 
This strategy can be adapted to yield an Hamming-accurate ranking by changing its stopping criterium.
Instead of stopping once all items have been eliminated, we stop when either $\setb - \h$ items have been assigned to the top, or $\numitems - \setb - \h$ items have been assigned to the bottom. 
While this strategy yields an Hamming accurate ranking, its sample complexity is, up to logarithmic factors, equal to 
$
\sum_{i=1}^{k-\h} \frac{1}{ (\score_i - \score_{k+1})^{2}} +
\sum_{i=k+1+\h}^\numitems \frac{1}{ (\score_k - \score_i)^{2}} 
+ 2\h (\score_{\setb-\h} -  \score_{\setb+1+\h} )
$, 
which is strictly smaller than that of the Hamming-LUCB algorithm. 
As Figure~\ref{fig:cmp} shows, this strategy requires significantly more comparisons for finding an approximate ranking, thereby validating the benefits of our approach.

\section{Proofs}

In this section, we provide the proofs of our theorems.  In order to simplify notation, we assume without loss of generality (re-indexing as needed)
that the underlying permutation $\pi$
equal to the identity, so that $\score_{1} > \score_{2} > \ldots >
\score_{\numitems}$.

\subsection{Proof of Theorem~\ref{thm:LUCBHAMsuff}}
Our analysis uses an argument inspired by the proof  
of the performance guarantee of the original LUCB algorithm from the bandit literature, presented in~\cite{kalyanakrishnan_pac_2012}. 
We begin by showing that the estimate
$\scorehat_i(\NC_i)$ is guaranteed to be $\alpha_i$-close to
$\score_i$, for all $i$, with high probability.  

\begin{lemma}[{\cite[Lem.~19]{kaufmann_complexity_2014}}]
\label{lem:probound}
For any $\delta \in (0,0.0005)$, with probability at least $1 - \delta$, the event 
\begin{align}
\label{eq:taubounded}
\eventscore \defeq \{ \left| \scorehat_i(\timeind) - \score_i \right|
\leq \alpha_i, \quad \text{for all $i \in [\numitems]$ and for all
  $\timeind \geq 1$} \}
\end{align}
occurs. 
The statement continues to hold for any $\delta \in (0,1)$ with 
$\alpha_i = \alpha(T_i) = \sqrt{ \frac{\beta(T_i,\delta')}{2T_i} }$, 
$\beta(\timeind,\delta') = 2 \log(125 \log(1.12 \timeind)/\delta')$. 
\end{lemma}

Lemma~\ref{lem:probound} is a non-asymptotic version of the law
of the iterated logarithm from \citet{kaufmann_complexity_2014} and \citet{jamieson_lil_2014}.

We first show that, on the event $\eventscore$ defined in equation~\eqref{eq:taubounded}, the 
Hamming-LUCB algorithm returns sets $\Shat_1$ and $\Shat_2$ obeying $\HD(\Shat_\setind, \S_\setind) \leq 2\h \text{ for } \ell =1,2$, as desired. 
Indeed, suppose that $\{(1),\ldots, (\setb-\h)\} \subseteq \S_1$. 
This implies that $\S_1$ and $\Shat_1$ differ in at most $\h$ values, which in turn implies 
that $\S_2$ and $\Shat_2$ differ by at most $\h$ values. 
Therefore, $\HD(\Shat_\setind, \S_\setind) \leq 2\h \text{ for } \ell =1,2$. 
Next, suppose that $\{(1),\ldots, (\setb-\h)\} \nsubseteq \S_1$. Then, 
at least one item in $\{(1),\ldots, (\setb-\h)\}$ is in $\S_2$. 
Thus, 
on $\eventscore$,  the termination condition~\eqref{eq:termination} implies that $\{(\setb +1 +\h), \ldots, (\numitems) \} \subset \{\setb+1, \ldots, \numitems \} = \S_2$. 
Similarly as above, this in
turn implies that $\HD(\Shat_\setind, \S_\setind) \leq 2\h \text{ for } \ell =1,2$.

We next show that on the event $\eventscore$, Hamming-LUCB terminates after the desired number of comparisons. 
Let $\gamma \defeq \frac{\score_{\setb - \h} + \score_{\setb + 1 + \h}}{2}$, and define the event that item $i$ is bad as 
\[
\bad(i) = 
\begin{cases}
\scorehat_i  < \gamma + 3\alpha_i, & i \in \{1,\ldots, \setb - \h\} \\
\scorehat_i  > \gamma - 3\alpha_i, & i \in \{\setb + 1 + \h,\ldots, \numitems \} \\
\alpha_i  > \frac{\score_{\setb - \h} - \score_{\setb + 1 + \h}}{4}, & \text{otherwise}. \\
\end{cases}
\]

\begin{lemma}
\label{lem:oneisbad}
If $\eventscore$ occurs and the termination condition~\eqref{eq:termination} is false, 
then either $\bad(\blow_1)$ or $\bad(\bup_2)$ occurs. 
\end{lemma}

Given Lemma~\ref{lem:oneisbad}, we can complete the proof in the following way. 
For an item $i$, define 
\[
\Delta_i 
=
\begin{cases}
\score_i - \score_{\setb+1+\h}, 
& i \in \{1, \ldots, \setb-\h \}  \\
\score_{\setb-\h} - \score_i 
&  i \in \{ \setb + 1 + \h, \ldots, \numitems \} \\
\score_{\setb - \h} - \score_{\setb+1+\h},
& \text{ otherwise}, 
\end{cases}
\]
and let $\NCb_{i}$ be the largest integer $u$ 
satisfying the bound
$
\alpha(u)
\leq \Delta_i/4
$. 
A simple calculation (see Section~\ref{sec:simplecomp} for the details) yields that
\begin{align}
\label{eq:factnuneedy}
\text{On the event $\eventscore $, if $\NC_i \geq \NCb_{i}$ holds, then $\bad(i)$ is false}.
\end{align}
Let $\timeind \geq 1$ be the $\timeind$-th iteration of the steps in the LUCB algorithm, and
let $\blow_1$ and $\bup_2$ be the two items selected in Step~\ref{it:step4} of the algorithm. Note that in each iteration only those two items are compared to other items. 
By Lemma~\ref{lem:oneisbad}, we can therefore bound the total number comparisons by 
\begin{align}
2\sum_{t=1}^{\infty} 
\ind{
\bad(\blow_1)
\cup
\bad(\bup_2)
} 
&\leq
2\sum_{\timeind=1}^{\infty} 
\sum_{i=1}^\numitems
\ind{
(i = \blow_1 \cup i = \bup_2)
\cap \bad(i)
} \nonumber \\
&\mystackrel{(i)}{\leq}
2
\sum_{\timeind=1}^{\infty}
\sum_{i=1}^\numitems
\ind{
(i = \blow_1 \cup i = \bup_2)
\cap \NC_i \leq \NCb_{i}
} \nonumber \\
&\mystackrel{(ii)}{\leq}
2
\sum_{i=1}^{\numitems}
\NCb_{i}.
\label{eq:indtit}
\end{align}
For inequality~(i), we used the fact~\eqref{eq:factnuneedy},
and inequality~(ii) follows because 
$\NC_{i}(t) \leq \NCb_{i}$ can only be true for 
$\NCb_{i}$ iterations $\timeind$. 

We conclude the proof by noting that the definition of $\alpha(\cdot)$ and some algebra yields (see~\cite[Eq.~(20)]{heckel_active_2016}) that for $\constone$ sufficiently large
\begin{align*}
\NCb_{i}
\leq \frac{\constone}{ (\Delta_i/4)^2 } \log\left(
\frac{n}{\delta} \log\left(\frac{2}{(\Delta_i/4)^2} \right)
\right)
\leq
c_2
\log\left(\frac{\numitems}{\delta}\right)
\frac{\log(2\log(2/ \Delta_i  ))}{ \Delta_i^2 }.
\end{align*}
Applying this inequality to the RHS of equation~\eqref{eq:indtit} above concludes the proof. 


\subsubsection{\label{sec:simplecomp}Proof of fact~\eqref{eq:factnuneedy}}

First, consider an item $i \in \{ \setb + 1 + \h, \ldots, \numitems \}$. 
We show that if $\NC_i \geq \NCb_i$, then $\bad(i)$ is false. 
On the event $\eventscore$, 
\begin{align}
\scorehat_i(\NCb_{i})
+
\alpha(\NCb_{i})
\leq 
\score_i + 2 \alpha(\NCb_{i})
\mystackrel{(i)}{\leq}
\score_i + \frac{ \Delta_i }{2}
= 
\gamma
+
\frac{\Delta_i}{2}
-
\frac{ \score_{\setb -\h} - \score_i  + \score_{\setb + 1 + \h}  - \score_i }{2}
\leq \gamma,
\end{align}
where inequality~(i) follows from 
$\alpha(\NC_i) \leq \Delta_i/4$ 
for $\NC_i \geq \NCb_{i}$, by definition of $\NCb_{i}$, and the last inequality follows from 
$\Delta_i = \score_{\setb-\h} - \score_i$
and $\score_{\setb+1+\h} - \score_i \geq 0$. 
Thus, $\bad(i)$ does not occur. 

For an item $i \in \{1,\ldots, \setb - \h\}$, $\bad(i)$ that is false, the argument is equivalent. 
For an item in the middle 
$i \in \{\setb -\h + 1, \ldots, \setb + \h\}$, the event $\bad(i)$ is false by definition.
This concludes the proof.


\subsubsection{Proof of Lemma~\ref{lem:oneisbad}}

We prove the lemma by considering all different values 
the indices $\blow_1$ and $\bup_2$ selected by the LUCB algorithm can take on, and showing that in each case $\bad(\blow_1)$ and $\bad(\bup_2)$ cannot occur simultaneously. 
For notational convenience, we define the indices
\[
\mind_1 = \underset{i \in \{ (\setb-\h+1), \ldots, (\setb) \} }{\arg \max}  \alpha_i,
\quad 
\mind_2 = \underset{i \in \{ (\setb+1), \ldots, (\setb+\h) \} }{ \arg \max } \alpha_i,
\]
and note that 
\begin{align*}
\blow_1 = \underset{ i \in \{ \dind_1,\mind_1\} }{\arg \max} ~ \alpha_i,
\quad
\text{and}
\quad 
\bup_2 = \underset{i \in \{\dind_2,\mind_2\}}{\arg \max} ~ \alpha_i. 
\end{align*}

\begin{enumerate}

\item Suppose that 
$\blow_1 \in \{1,\ldots, \setb - \h\}$ 
and 
$\bup_2 \in \{\setb + 1 + \h,\ldots, \numitems\}$, 
and that both 
$\bad(\blow_1)$ 
and
$\bad(\bup_2)$ do not occur. 
First note that 
\begin{align}
\scorehat_{\dind_1} - \alpha_{\dind_1} 
\geq
\scorehat_{\blow_1} - \alpha_{\blow_1}.
\end{align}
In order to establish this claim, note that the inequality holds trivially with equality if $\blow_1 = \dind_1$. 
If $\blow_1 = \mind_1$, then 
it follows from 
$\scorehat_{\dind_1} \geq \scorehat_{ \mind_1 }$ 
and $\alpha_{\dind_1} \leq \alpha_{\blow_1}$. 
Thus, we obtain
\begin{align}
\label{eq:schatdblowg}
\scorehat_{\dind_1} - \alpha_{\dind_1} 
\geq
\scorehat_{\blow_1} - \alpha_{\blow_1}
> \gamma,
\end{align}
where the last inequality holds by the assumption that $\bad(\blow_1)$ does not occur. 
An analogous argument yields that
\begin{align}
\label{eq:lownotermcon}
\gamma 
>
\scorehat_{\bup_2} + \alpha_{\bup_2}
\geq
\scorehat_{\dind_2} + \alpha_{\dind_2}.
\end{align}
Combining those inequalities yields 
$\scorehat_{\dind_1} - \alpha_{\dind_1} > \scorehat_{\dind_2} + \alpha_{\dind_2}$, which contradicts that the termination condition~\eqref{eq:termination} is false. 


\item Next, suppose that $\blow_1$ is an index in the middle and 
$\bup_2$ is in the very bottom, i.e., $\blow_1 \in \{\setb - \h+1, \ldots, \setb+\h\}$, 
and
$\bup_2 \in \{\setb + 1 + \h,\ldots, \numitems\}$, 
and both $\bad(\blow_1)$ and $\bad(\bup_2)$ do not occur. 

First note that from $\bup_2 \in \{\setb + 1 + \h,\ldots, \numitems\}$ and
$\bad(\bup_2)$ not occurring, we have that
\begin{align*}
\gamma \geq \scorehat_{\bup_2} + 3\alpha_{\bup_2}
&\mystackrel{(i)}{\geq}
\scorehat_{\dind_2} + \alpha_{\dind_2} + 2\alpha_{\bup_2} 
\mystackrel{(ii)}{\geq}
\scorehat_{i} + \alpha_{i} + 2\alpha_{\bup_2},
\quad
\text{
for $i \in \{(\setb+1+\h),\ldots,(\numitems)\}$
}.
\end{align*}
Here, inequality~(i) holds by 
$\scorehat_{\bup_2} \geq \scorehat_{\dind_2}$ 
and $\alpha_{\bup_2} \geq \alpha_{\dind_2}$, 
and inequality~(ii) follows by the definition of $\dind_2$.
On the event $\eventscore$, this implies 
\begin{align}
\gamma 
\geq 
\score_i + 2\alpha_{\bup_2}.
\label{eq:gamlier2}
\end{align}
Inequality~\eqref{eq:gamlier2} can only be true for all $i \in \{(\setb+1+\h),\ldots,(\numitems)\}$
if
$ \gamma - \score_{k+1+\h} \geq 2\alpha_{\bup_2} $, which is equivalent to 
\[
\alpha_{\bup_2} \leq \frac{\Delta}{4}, \quad  \Delta \defeq \score_{k - \h} - \score_{k+1+\h}.
\]

Again using that $\bup_2 \in \{\setb + 1 + \h,\ldots, \numitems\}$ and 
$\bad(\bup_2)$ not occurring, we have that
\begin{align}
\label{eq:gamdeal1}
\gamma
\geq
\scorehat_{\dind_2} + \alpha_{\dind_2} 
\mystackrel{(i)}{\geq} 
\scorehat_{\dind_1} - \alpha_{\dind_1}
\mystackrel{(ii)}{\geq}
\scorehat_{\dind_1} - \frac{\Delta}{4} ,
\end{align}
where inequality~(i) holds since the termination condition~\eqref{eq:termination} is false, and inequality~(ii) follows from 
$\alpha_{\dind_1} \leq \alpha_{\blow_1} \leq \frac{\Delta}{4}$, 
where the last inequality holds since $\bad(\bup_2)$ does not occur, by assumption. 

From $\scorehat_{\dind_1} \geq \scorehat_i$ for all $i \in \{(k-\h+1), \ldots, (\numitems)\}$, 
it follows that for $i \in \{\dind_1\}\cup\{(k-\h+1), \ldots, (\numitems)\}$,
\begin{align}
\gamma > \scorehat_{i} - \frac{\Delta}{4}
\geq
\score_i - \alpha_i - \frac{\Delta}{4}. 
\end{align}
Below, we show that 
\begin{align}
\label{eq:middlesmall}
\text{$\alpha_i \leq \frac{\Delta}{4}$, \quad for all $i \in 
\{\dind_1\} \cup \{ (k-\h+1) ,\ldots, (k+\h) \}$.}
\end{align}
It follows that
\begin{align}
\label{eq:asciadf}
\gamma > \score_i - \frac{\Delta}{4} - \frac{\Delta}{4} \Leftrightarrow
\score_{k - \h} > \score_i , \quad \text{ for all } i \in \{\dind_1\} \cup \{(k-\h+1), \ldots, (k+\h) \}.
\end{align}
Together with equation~\eqref{eq:gamlier2}, this yields that 
$\score_{k - \h} > \score_i$ for all $i \in \{\dind_1\} \cup \{(k-\h+1), \ldots, (\numitems) \}$, 
which is a contradiction. This concludes the proof. 

It remains to establish the claim~\eqref{eq:middlesmall}. 
From the bound $\alpha_{\bup_2} \leq \frac{\Delta}{4}$, as shown above, we have 
$\frac{\Delta}{4} \geq \alpha_{\bup_2} \geq \alpha_{\mind_2} \geq \alpha_i$ for all $i \in \{(k+1),\ldots, (k+\h)\}$, by defintion of $\mind_2$. 
Moreover, for $i \in \{(k-\h), \ldots, (k)\}$, we have $\alpha_i \leq \alpha_{\dind_1} \leq \alpha_{\blow_1} \leq \frac{\Delta}{4}$, where the last inequality holds since $\blow_1$ is in the middle and is not bad. This concludes the proof of~\eqref{eq:middlesmall}. 


\item 
The case where $\blow_1$ lies in the very top and $\bup_2$ lies in the middle, i.e., 
$\blow_1 \in \{1,\ldots, k - \h\}$
and
$\bup_2 \in \{\setb - \h+1, \ldots, \setb+\h\}$, 
and both $\bad(\blow_1)$ and 
$\bad(\bup_2)$ do not occur, 
can be treated analogously as the previous case.

\item 
Next, suppose that both $\blow_1$ and $\bup_2$ lie in the middle, i.e., $\blow_1, \bup_2 \in \{\setb - \h+1, \ldots, \setb+\h\}$
and both $\bad(\blow_1)$ and $\bad(\bup_2)$ do not occur. 
We show that this leads a contradiction. 

Towards this goal, first note that 
either 
\begin{align}
\label{eq:middleassst}
\gamma < \scorehat_{\dind_2} + \alpha_{\dind_2}.  
\end{align}
holds true or
\begin{align}
\label{eq:middleassst2}
\gamma
>
\scorehat_{\dind_1} - \alpha_{\dind_1},
\end{align}
holds true, but not both. 
In order to see this fact,
note that if inequality~\eqref{eq:middleassst} is violated, then 
\begin{align}
\gamma \geq 
\scorehat_{\dind_2} + \alpha_{\dind_2} 
\mystackrel{(i)}{<}
\scorehat_{\dind_1} - \alpha_{\dind_1},
\label{eq:gamleqdind1}
\end{align}
where step~(i) follows from the termination condition~\eqref{eq:termination} being false, by assumption. 
Likewise, if inequality~\eqref{eq:middleassst2} does not hold, then
\[
\gamma
\leq
\scorehat_{\dind_1} - \alpha_{\dind_1} < \scorehat_{\dind_2} + \alpha_{\dind_2}. 
\]

We have shown that either condition~\eqref{eq:middleassst} or~\eqref{eq:middleassst} holds true, but not both simultaneously; consequently, we may conclude that at least one of these two conditions does not hold true. 
Next, we show that if either inequality~\eqref{eq:middleassst} or inequality~\eqref{eq:middleassst} does not hold true, then this leads to a contradiction, which concludes the proof. 

First, suppose that inequality~\eqref{eq:middleassst} \emph{does not} hold true. 
Then, by definition of $\dind_2$, on $\eventscore$, 
\begin{align}
\label{eq:fierstcontt}
\gamma 
\geq 
\scorehat_{\dind_2} + \alpha_{\dind_2}
\geq 
\scorehat_{i} + \alpha_{i}
\geq
\score_i,
\quad
\text{for all $i \in \{(k+1+\h), \ldots, (\numitems) \}
$}. 
\end{align}
Moreover, by inequality~\eqref{eq:gamleqdind1} together with $\bad(\blow_1)$ and $\bad(\bup_2)$ not occurring, which implies that $\alpha_{\blow_1},\alpha_{\bup_2} \leq \frac{\Delta}{4}$, 
the following inequality follows by the same argument as inequality~\eqref{eq:asciadf} follow from inequality~\eqref{eq:gamdeal1}:
\begin{align}
\score_{k - \h} > \score_i,
\quad
\text{ for all } 
i \in \{\dind_1\} \cup \{(k-\h+1), \ldots, (k+\h) \}.
\end{align}
Together with condition~\eqref{eq:fierstcontt}, this yields a contradiction. 

The argument for the case in which claim~\eqref{eq:middleassst} is true is entirely analogous. 

\item Finally, if 
$\blow_1 \in \{\setb + 1 + \h,\ldots, \numitems\}$
or if 
$\bup_2 \in \{1,\ldots, \setb - \h\}$,
and both $\bad(\blow_1)$ and 
$\bad(\bup_2)$ do not occur,
we reach a contradiction using similar arguments as in the previous cases. 
\end{enumerate}




\newcommand\nuone{\nu}
\newcommand\nutwo{\nu'}

\newcommand\eventone{W}
\newcommand\eventtwo{W'}


\subsection{Proof of Theorem~\ref{thm:lowerbound}}

We now turn to the proof of the lower bound from
Theorem~\ref{thm:lowerbound}. 

We first introduce some notation required
to state a useful lemma~\cite[Lem.~1]{kaufmann_complexity_2014} from
the bandit literature. 
Let $\nu = \{\nu_j\}_{j=1}^m$ be a collection
of $m$ probability distributions, each supported on the real line
$\reals$.  Consider an algorithm $\alg$, that, at times
$\timeind=1,2,\ldots$, selects the index $i_\timeind \in [m]$ and
receives an independent draw $\bernrv_\timeind$ from the distribution
$\nu_{i_\timeind}$ in response.  Algorithm $\alg$ may select
$i_\timeind$ only based on past observations, that is,
$i_\timeind$ is $\mc F_{\timeind-1}$ measurable, where
$\sigmaalgebra_\timeind$ is the $\sigma$-algebra generated by
$i_1,\bernrv_{i_1},\ldots,i_\timeind,\bernrv_{i_\timeind}$.
Algorithm $\alg$ has a stopping rule $\stoptime$ that determines the
termination of $\alg$.  We assume that $\stoptime$ is a stopping time
measurable with respect to $\sigmaalgebra_\timeind$ and obeying
$\PR{\stoptime < \infty} =1$.

Let $\numcmp_i(\stoptime)$ denote the total number of times index $i$
has been selected by the algorithm $\alg$ (until termination).  For
any pair of distributions $\nu$ and $\nu'$, we let $\KL(\nu,
\nu')$ denote their Kullback-Leibler divergence, and for any $p, q \in
    [0,1]$, let $\kl(p,q) \defeq p \log \frac{p}{q} + (1-p) \log
    \frac{1-p}{1-q}$ denote the Kullback-Leiber divergence between two
    binary random variables with success probabilities $p, q$.

With this notation, the following lemma relates the cumulative number
of comparisons to the uncertainty between the actual distribution
$\nu$ and an alternative distribution $\nu'$.

\begin{lemma}[{\cite[Lem.~1]{kaufmann_complexity_2014}}]
Let $\nu,\nu'$ be two collections of $m$ probability distributions on
$\reals$.  Then for any $\event \in \sigmaalgebra_\stoptime$ with
$\PR[\nu]{\event} \in (0,1)$, we have
\begin{align}
\sum_{i=1}^m \EX[\nu]{\numcmp_i(\stoptime)} \KL(\nu_i,
\nu_i') \geq \kl(\PR[\nu]{\event}, \PR[\nu']{\event}).
\end{align}
\label{lem:changemeasure}
\end{lemma}

\noindent Let us now use Lemma~\ref{lem:changemeasure} to prove
Theorem~\ref{thm:lowerbound}. 

Define the event 
\[
\event \defeq \left\{  \HD(\Shat_\setind, \S_\setind) \leq 2\h \text{ for } \ell =1,2  \right\},
\]
corresponding to success of the algorithm $\alg$. 
Recalling that
$\stoptime$ is the stopping rule of algorithm $\alg$, we are
guaranteed that $\event \in \sigmaalgebra_\stoptime$.  Given the
linear relations $\pmat_{ij} = 1- \pmat_{ji}$, the pairwise comparison
matrix $\Pmat$ is determined by the entries $\{ \pmat_{ij},
i=1,\ldots,\numitems,\; j = i+1,\ldots,\numitems \}$.  Let
$\numcmp_{ij}(\stoptime)$ be the total number of comparisons between
items $i$ and $j$ made by $\alg$.  For any other pairwise comparison
matrix $\Pmat' \in \comparisonclass[0]$, Lemma~\ref{lem:changemeasure}
ensures that
\begin{align}
\label{eq:lemspecialized}
\sum_{i=1}^\numitems \sum_{j=i+1}^\numitems \EX[\Pmat]{\numcmp_{ij}}
\kl(\pmat_{ij}, \pmat_{ij}') \geq \kl(\PR[\Pmat]{\event},
\PR[\Pmat']{\event}).
\end{align}

Let $\setM \defeq \{ \m_1,\ldots, \m_{2\h+1} \}$ be a set of distinct items in $\S_1$. 
We next construct 
$\Pmat' \in
\comparisonclass[1/8]$ such that 
$\m_1,\ldots, \m_{2\h+1} \notin \S_1(\Pmat')$ under the distribution $\Pmat'$. 
Since we assume algorithm $\alg$ to be uniformly $(\h,\delta)$-Hamming-accurate over $\comparisonclass[1/8]$, we have both
$\PR[\Pmat]{\event} \geq 1 - \delta$ and $\PR[\Pmat']{\event} \leq
\delta$. 
To see this note that since $\S_1$ and $\S_1(\Pmat')$ differ in $2\h +1$ elements, there is no set of cardinality $\setb$ that differs from both $\S_1$ and $\S_1(\Pmat')$ in only $\h$ elements. It follows that
\begin{align}
\kl(\PR[\Pmat]{\event}, \PR[\Pmat']{\event}) \geq \kl(\delta,1-\delta)
= (1-2\delta) \log \frac{1-\delta}{\delta} \geq \log \frac{1}{2\delta},
\label{eq:lbdApl2}
\end{align}
where the last inequality holds for $\delta \leq 0.15$. 

It remains to specify the alternative matrix $\Pmat' \in \comparisonclass[0]$. 
The alternative matrix $\Pmat'$ is defined as 
\begin{align}
\label{eq:pdashijHam}
\pmat_{ij}' =
\begin{cases}
\pmat_{\m j} - 
\frac{\numitems - 1}{\numitems - 1 - 2\h}
(\score_{\m} - \score_{\setb+1+2\h}) , 
& \text{if } i = \m \text{ for } \m \in \setM, j \in [\numitems] \setminus \setM \\ 
\pmat_{i \m} +
\frac{\numitems - 1}{\numitems - 1 - 2\h} (\score_{\m} - \score_{\setb+1+2\h}) , & \text{if } j = \m \text{ for } \m \in \setM, i \in
[\numitems] \setminus \setM \\ \pmat_{ij} & \text{otherwise}.
\end{cases}
\end{align}
It follows that, for $\m \in \setM$,
\begin{align*}
\score_\m' 
&= 
\frac{1}{\numitems-1} \sum_{j \in [\numitems] \setminus
  \{\m\} } \pmat_{\m j}' \\
&= \frac{1}{\numitems-1}\sum_{j \in
  [\numitems] \setminus \{\m\}} \pmat_{\m j}
  -
 \frac{1}{\numitems-1}
 \sum_{j \in [\numitems] \setminus \setM }
  \frac{\numitems - 1}{\numitems - 1 - 2\h}
(\score_{\m} - \score_{\setb+1+2\h})  \\
&= 
\score_{ \setb + 1 + 2\h }.
\end{align*}
Similarly, all other scores $\score_i'$ are larger than $\score_i$ by
a common constant, that is, for \mbox{$i \in [\numitems] \setminus \setM$}, 
\begin{align*}
\score_i' = 
\score_i + \frac{1}{\numitems - 1 - 2\h} \sum_{\m \in \setM} (\score_{\m} - \score_{\setb+1+2\h}).
\end{align*}
It follows that, under the distribution $\Pmat'$ the items in the set $\setM$ are not among the $\setb$ highest scoring items, which ensures that $\setM \cap \S_1(\pmat) = \emptyset$. 
Moreover, $\Pmat' \in \comparisonclass[1/8]$.
This follows from the assumption $\Pmat \in \comparisonclass[3/8]$,
which implies
\begin{align*}
\pmat_{\m j}' \leq \frac{5}{8} + \left(\frac{5}{8} - \frac{3}{8} \right)  \leq \frac{7}{8}, 
\end{align*}
and similarity $\pmat_{\m j}' \geq \frac{1}{8}$.

Next consider the total number of comparisons of item $\m$ with all
others items,\linebreak \mbox{that is, $ \numcmp_\m = \sum_{ j \in [\numitems]
    \setminus \{\m\} } \numcmp_{\m j}$.} 
By linearity of expectation, we have
\begin{align}
\sum_{\m \in \setM}
\max_{j \in [\numitems] \setminus \{\m \}} \kl(\pmat_{\m j}, \pmat_{\m
  j}') \EX[\Pmat]{\numcmp_\m} 
  &= \sum_{\m \in \setM} \max_{j \in [\numitems] \setminus \{
  \m\}} \kl(\pmat_{\m j}, \pmat_{\m j}') \sum_{j' \in [\numitems]
  \setminus \{\m\} } \EX[\Pmat]{\numcmp_{\m j'}} \nonumber \\
&\mystackrel{(i)}{\geq} 
\sum_{\m \in \setM}
\sum_{j \in [\numitems]\setminus\{\m \}} \EX[\Pmat]{\numcmp_{\m
    j}} \kl(\pmat_{\m j}, \pmat_{\m j}') \nonumber \\
& \mystackrel{(ii)}{=} \sum_{i=1}^\numitems \sum_{j=i+1}^\numitems
\EX[\Pmat]{\numcmp_{ij}} \kl(\pmat_{ij}, \pmat_{ij}') \nonumber \\
& \mystackrel{(iii)}{\geq} \kl(\PR[\Pmat]{\event}, \PR[\Pmat']{\event})
\nonumber \\
\label{eq:bylbdApl2}
& \geq \log \frac{1}{2\delta}.
\end{align}
Here steps~(i) and (ii) follows from the fact that $\kl(\pmat_{ij},
\pmat_{ij}')=0$ for all $(i,j)$ not in 
$\{(\m,j) \mid \m \in \setM, j \in [\numitems] \setminus \setM \}$ 
and not in 
$\{ (i,\m) \mid \m \in \setM, i \in [\numitems] \setminus \setM \}$, 
by definition of the $\pmat'_{ij}$ (see equation~\eqref{eq:pdashijHam}), and step~(iii)
     follows from inequality~\eqref{eq:lemspecialized} (that is, from
     Lemma~\ref{lem:changemeasure}). Finally,
     inequality~\eqref{eq:bylbdApl2} follows from
     inequality~\eqref{eq:lbdApl2}.

We next upper bound the KL divergence on the left hand side of
inequality~\eqref{eq:bylbdApl2}.
Using the inequality $\log x \leq
x-1$ valid for $x>0$, we have that 
\begin{align}
\kl(\pmat_{\m j}, \pmat_{\m j}')
&\leq \frac{(\pmat_{\m j} - \pmat_{\m j}')^2}{\pmat_{\m j}'
  (1-\pmat_{\m j}') }  \; 
  \leq 
  d_{\m} , \quad d_{\m} \defeq 16 (\score_\m - \score_{\setb+1+2\h})^2.
\label{eq:klbscore2}
\end{align}
Here, the last inequality follows from the definition of
$\Pmat'$ in equation~\eqref{eq:pdashijHam}, for $j \in [\numitems]
\setminus \{\m \}$, and from $\frac{1}{8} \leq \pmat_{\m j}' \leq
\frac{7}{8}$, which implies $\frac{1}{\pmat_{\m j}' (1-\pmat_{\m j}')
} \leq 16$. 
Applying inequality~\eqref{eq:klbscore2} to the left hand side of inequality~\eqref{eq:bylbdApl2} yields 
\begin{align}
\label{eq:constraints}
\sum_{\m \in \setM} d_\m  \EX[\Pmat]{\numcmp_\m}  
\geq 
\log \frac{1}{2\delta},
\quad \text{ valid for each subset $\setM \subseteq \S_1$ of cardinality $2\h+1$}.
\end{align}
We can therefore obtain a lower bound on 
$\sum_{i \in \S_1} \EX[\Pmat]{\numcmp_\m}$ 
by solving the minimization problem:
\begin{align}
&\underset{e_\m \geq 0}{\text{minimize}} \sum_{\m \in \S_1} e_\m \nonumber \\
&\text{subject to}
\quad
\sum_{\m \in \setM} d_\m e_\m
\geq \log \frac{1}{2\delta}
\quad
\text{
for each subset $\setM \subseteq \S_1$ of cardinality $2\h+1$}.
\end{align}
Since the $d_\m$ are decreasing in $\m$, the solution to this optimization problem is $e_{\setb - 2\h},\ldots,e_{\setb} = 0$ and 
$e_{\m} = \log(1/2\delta)/d_\m$. 

Using an analogous line of arguments for items in the set $\S_2$, we arrive at the following lower bound
\[
\log \frac{1}{2\delta}
\left(
\sum_{i = 1}^{\setb - 2\h} 
\frac{1}{8( \score_i  - \score_{\setb+1+2\h} )}
+ 
\sum_{i = \setb+1+2\h}^{\numitems}
\frac{1}{8( \score_{\setb - 2\h}  - \score_{i} )}
\right)
\]
on the number of comparisons. 
This concludes the proof.

%


\subsection{\label{sec:altcmp}Alternative lower bound}

In this section, we state a second lower bound on the number of
comparisons, which shows that to obtain an $(\h,\delta)$-Hamming
accurate ranking, an algorithm has to compare \emph{each} item a
certain number of times.  The proof of this lower bound also forms the
foundation for the proof of Theorem~\ref{thm:parametric}.

\begin{theorem} \label{thm:lowerboundmoderateconf}
Let $\alg$ be a symmetric algorithm, i.e., its distribution of comparisons commutes with permutations of the items, 
that is uniformly $(\h,\delta)$-Hamming accurate over $\comparisonclass$, with 
$\delta \leq \frac{1}{2} \min(\frac{1}{\setb}, \frac{1}{\numitems-\setb})$.
Choose an integer $q\geq 1$. 
Then, for any item $\it \in [\numitems]$, when applied to a given pairwise comparison model $\Pmat \in \comparisonclass$, the algorithm $\alg$ must make at least 
\begin{align*}
\frac{2}{3}  \left( \frac{2q -1}{2\h + q} \right)^2
\Big/
\left(
\max_{\alt \in 
\{\setb - 2(\h+q), \ldots, \setb + 1 +  2(\h + q)\} 
}
\max 
\left(
\max_{ j \neq \{\it,\alt\} }
\kl(\pmat_{\it j}, \pmat_{\alt j})
,
\kl(\pmat_{\it \alt}, 1/2)
\right)
\right) 
\end{align*}
comparisons on average. 
\end{theorem}

In the remainder of this section, we provide a proof of
Theorem~\ref{thm:lowerboundmoderateconf}.  For a given item $\it$, we
divide our proof into two cases, corresponding to whether or not $\PR{
  \it \notin \Shat_1 } > c_1 + \eta$, where we define the scalar $\eta
\defeq \frac{1 - c_1 - c_2}{2}$.

\paragraph{Case 1:}

First, suppose that $\PR{ \it \notin \Shat_1 } > c_1 + \eta$.  Pick
some other item $\alt$ in $\{\setb - \h', \ldots, \setb \}$ that obeys
$\PR{ \alt \notin \Shat_1 } \leq c_1$.  The following lemma guarantees
that such an item exists:
\begin{lemma}
\label{lem:alternatives}
Let $\alg$ be an algorithm that is $(\h,\delta)$-Hamming accurate,
with $\delta \leq \frac{1}{2} \min(\frac{1}{\setb},
\frac{1}{\numitems-\setb})$.  Let $\Shat_1$ and $\Shat_2$ be $\alg$'s
estimate of the top $\setb$ items $\S_1$ and the bottom $\numitems -
\setb$ items $\S_2$, respectively.  Choose constants $c_1,c_2$ and
$\h'$ such that $\h + \frac{1}{2} \leq c_1 h'$, $\h + \frac{1}{2} \leq
c_2 h'$, and $c_1+c_2 < 1$.  Then
\begin{itemize}
\item[i)] there exists an item $\alt \in \{\setb - \h', \ldots, \setb
  \}$ such that $\PR{ \alt \notin \Shat_1 } \leq c_1$, and
\item[ii)] there exists an item $\alttwo \in
  \{\setb+1,\dots,\setb+1+\h'\}$ such that $\PR{\alttwo \in
    \Shat_1}\leq c_2$.
\end{itemize}
\end{lemma}

We use Lemma~\ref{lem:changemeasure} from Kaufmann et al., which
relates the expected number of comparisons to the uncertainty between
the actual distribution $\pmat$ and an alternative distribution
$\pmat'$ about the events $\event_\it \defeq \{\it \notin \Shat_1\}$
and $\event_\alt = \{\alt \notin \Shat_1\}$.  Concretely, define the
alternative matrix $\pmat'$ as
\begin{align}
\label{eq:constraltmtx}
\pmat'_{ij} =
\begin{cases} 
\pmat_{\alt j}, & i = \it, j \in [\numitems] \setminus \{\it,\alt \}
\\ \pmat_{i \alt}, & j = \it , i \in [\numitems] \setminus \{\it,\alt
\} \\ 1/2, & i = \it \text{ and } j = \alt, \text{ or } i = \alt
\text{ and } j = \it \\ \pmat_{i,j}, & \text{ otherwise}.
\end{cases}
\end{align}
Since the algorithm $\alg$ is invariant to permutations of the labels,
by assumption, we have that $\PR[\pmat']{\event_\it} =
\PR[\pmat']{\event_\alt}$, since $\it$ and $\alt$ have the same
distribution under the distribution specified by $\pmat'$, and we
assume $\alg$ to be symmetric.  Moreover, by construction of $\pmat'$,
we have
\begin{align*}
\sum_{i=1}^\numitems \sum_{j = i+1}^\numitems
\EX[\nu]{\numcmp_{ij}(\stoptime)} \KL(\pmat_{ij},\pmat_{ij}') =
\sum_{j \notin \{\it,\alt\}} \EX[\pmat]{\numcmp_{\it j}(\stoptime)}
\KL(\pmat_{\it j},\pmat_{\alt j}) + \EX[\pmat]{\numcmp_{\it
    \alt}(\stoptime)} \KL(\pmat_{\it j}, 1/2).
\end{align*}
Applying Lemma~\ref{lem:changemeasure} from Kaufmann et al.  (see
equation~\eqref{eq:lemspecialized}) then yields
\begin{align}
\hspace{2cm}&\hspace{-2cm} \sum_{j \notin \{\it,\alt\}}
\EX[\pmat]{\numcmp_{\it j}(\stoptime)} \KL(\pmat_{\it j},\pmat_{\alt
  j}) + \EX[\pmat]{\numcmp_{\it \alt}(\stoptime)} \KL(\pmat_{\it j},
1/2) \nonumber \\
&\geq \max\{\kl(\PR[\nu]{\event_\it}, \PR[\nu']{\event_\it}),
\kl(\PR[\nu]{\event_\alt}, \PR[\nu']{\event_\alt})\} \nonumber \\
&= \max\{\kl(\PR[\nu]{\event_\it}, \PR[\nu']{\event_\it}),
\kl(\PR[\nu]{\event_\alt}, \PR[\nu']{\event_\it})\} \nonumber \\
&\geq \min_{p \in [0,1]}\max\{\kl(\PR[\nu]{\event_\it}, p),
\kl(\PR[\nu]{\event_\alt}, p)\} \nonumber \\
&\geq 
\frac{2}{3}  \eta^2,
\label{eq:Njaltone}
\end{align}
where the last inequality follows from Lemma~\ref{lem:klav} stated
below, together with $\PR{ \it \notin \Shat_1 } - \PR{ \alt \notin
  \Shat_1 } \geq \eta$, which follows from $\PR{ \it \notin \Shat_1 }
> c_1 + \eta$ and $\PR{ \alt \notin \Shat_1 } \leq c_1$.
\begin{lemma}
\label{lem:klav}
For scalar $p_a, p_b \in [0,1]$, let $\bar{p}$ denote their average
$\frac{p_a+p_b}{2}$. Then,
\begin{align}
\label{eq:minmaxdpap}
\kl(p_a,\bar{p}) + \kl(p_b,\bar{p})
\geq
\min_{p \in [0,1]}\max\{\kl(p_a, p), \kl(p_b, p)\}  
\geq
\frac{1}{2}(\kl(p_a,\bar{p}) + \kl(p_b,\bar{p})) 
.
\end{align}
Moreover, if $p_b - p_a \geq \eta$, then
\begin{align}
\label{eq:dpapleeta}
\frac{1}{2}(\kl(p_a,\bar{p}) + \kl(p_b,\bar{p})) 
\geq 
\frac{2}{3}  \eta^2.
\end{align}
\end{lemma}

\paragraph{Case 2:}  Turning to the other case,
suppose that $\PR{ \it \notin \Shat_1 } \leq c_1 + \eta$.  Pick some
other item $\alttwo$ in $\{\setb+1,\dots,\setb+1+\h'\}$ obeying
$\PR{\alttwo \notin \Shat_1} > 1 - c_2$, and note that
Lemma~\ref{lem:alternatives} ensures that such an item exits.  Using a
line of argument analogous to that above, we find that
\begin{align}
\label{eq:Njalttwo}
\sum_{j \notin \{\it,\alt\}} \EX[\pmat]{\numcmp_{\it j}(\stoptime)}
\KL(\pmat_{\it j},\pmat_{\alttwo j}) + \EX[\pmat]{\numcmp_{\it
    \alt}(\stoptime)} \KL(\pmat_{\it j}, 1/2)
& \geq \frac{2}{3} \eta^2.
\end{align}
Here we used that Lemma~\ref{lem:klav} together with the lower bound $
\PR{ \it \in \Shat_1 } - \PR{ \alttwo \in \Shat_1 } \geq (1 - c_1 -
\eta) - c_2 = \eta$, which in turn follows from the relations $\PR{
  \it \notin \Shat_1 } \leq c_1 + \eta$, $\PR{\alttwo \notin \Shat_1}
> 1 - c_2$, and $1 - c_1 - c_2 = 2\eta$.

Combining inequalities~\eqref{eq:Njaltone} and~\eqref{eq:Njalttwo}
yields
\begin{align}
\max_{\alt \in \{\setb - \h', \ldots, \setb + 1 + \h'\} } \left\{
\sum_{j \notin \{\it,\alt\}} \EX[\pmat]{\numcmp_{\it j}(\stoptime)}
\KL(\pmat_{\it j},\pmat_{\alt j}) + \EX[\pmat]{\numcmp_{\it
    \alt}(\stoptime)} \KL(\pmat_{\it j}, 1/2) \right\}
& \geq \frac{2}{3} \eta^2.
\end{align}
Choosing $\h'= 2(\h + q)$ and $c_1=c_2 = 1/2 - \eta$ concludes the proof. 



\subsubsection{Proof of Lemma~\ref{lem:alternatives}}

Since $\alg$ is $(\h,\delta)$-accurate, we have
\[
\sum_{i \in \S_1} \PR{ i \notin \Shat }
\leq 
\h + \delta \setb \leq \h + \frac{1}{2}
\leq 
c_1 \h',
\]
where the last inequality holds by assumption. 
Thus, there are at most $\h'$ many $i \in \S_1$ with $\PR{ i \notin \Shat } \geq c_1$, 
which implies that for at least $\setb - \h'$ many items $i \in \S_1$, we have that $\PR{ i \notin \Shat } \leq c_1$. 
This in turn implies that there is at least one item $\alt \in \{\setb - \h', \ldots, \setb \}$ obeying $\PR{ \alt \notin \Shat }
\leq c_1$.

Likewise, assuming that $\alg$ is $(\h,\delta)$-accurate, we have 
\[
\sum_{i \in \S_2} \PR{ i \in \Shat }
\leq \h + \delta (\numitems - \setb) 
\leq \h + \frac{1}{2} 
\leq c_2 \h'.
\]
Then there exists at least one arm
$\alttwo \in \{\setb+1,\dots,\setb+1+\h'\}$ such that $\PR{\alttwo \in \Shat_1}
\leq c_2$.


\subsubsection{Proof of Lemma~\ref{lem:klav}}

We start with proving inequality~\eqref{eq:minmaxdpap}. 
Observe that, since $\kl(x,y) \ge 0$, we have
\begin{align*}
\min_{p \in [0,1]}(\kl(p_a,p) + \kl(p_b,p)
\geq
\min_{p \in [0,1]}\max\{\kl(p_a, p), \kl(p_b, p)\}  
\geq
\min_{p \in [0,1]} \frac{1}{2}(\kl(p_a,p) + \kl(p_b,p)).
\end{align*}
Hence, it suffices to show that $\min_{p \in [0,1]} \frac{1}{2}(\kl(p_a,p) + \kl(p_b,p)) = \frac{1}{2}(\kl(p_a,\bar{p}) + \kl(p_b,\bar{p}))$. 
To this end, define the binary entropy 
$\ent(q) \defeq -q\log q - (1-q)\log (1-q)$. We then have
\begin{align*}
\frac{1}{2}(\kl(p_a, p) + \kl(p_b, p)) 
&= -\frac{1}{2}(\ent(p_a) + \ent(p_b)) +  \bar{p}\log \frac{1}{p} + (1-\bar{p})\log \frac{1}{1-p}\\
&= -\frac{1}{2}(\ent(p_a) + \ent(p_b)) + \ent(\bar{p}) +  \bar{p}\log\frac{\bar{p}}{p} + (1-\bar{p})\log\left(\frac{1-\bar{p}}{1-p}\right)\\
&= -\frac{1}{2}(\ent(p_a) + \ent(p_b)) + \ent(\bar{p})  + \kl(\bar{p},p),
\end{align*}
which is minimized by taking $p = \bar{p}$, for which $\kl(\bar{p},\bar{p}) = 0$. 
We can then expand
\begin{align*}
 -\ent(\bar{p}) 
 &= \frac{p_a+p_b}{2}\log(\bar{p}) + \left(1 - \frac{p_a+p_b}{2}\right) \log \left(1 - \bar{p}\right) \\
 &= \frac{p_a}{2} \log(\bar{p}) + \frac{1-p_a}{2} \log(1-\bar{p}) + \frac{p_b}{2} \log (\bar{p}) + \frac{1-p_b}{2}\log(1 - \bar{p}).
\end{align*}
Thus
\begin{align*}
-\frac{1}{2}(\ent(p_a) + \ent(p_b)) + \ent(\bar{p}) 
&= \frac{1}{2}(-\ent(p_a) - p_a\log(\bar{p}) - (1 - p_a)\log (1-\bar{p})) \\
&+\frac{1}{2}(-\ent(p_b) - p_b\log(\bar{p}) - (1-p_b)\log (1 - \bar{p})) \\
&=\frac{1}{2}\{\kl(p_a,\bar{p})+\kl(p_b,\bar{p})),
\end{align*}
as needed.

We next prove inequality~\eqref{eq:dpapleeta}. We have
\begin{align*}
\kl(p_a,\bar{p})
&=
\kl\left(p_a,\frac{p_b - p_a}{2} + p_a \right) \nonumber \\
&\geq
\min_{p\in [0,1]}
\kl(p, \eta/2 + p)
= 
\kl(1/2 + \eta/4, 1/2-\eta/4) 
=
\frac{\eta}{2} \log \left( \frac{1/2 + \eta/4}{1/2 - \eta/4} \right)  \\
&\geq \frac{\eta}{2} \left( 1 - \frac{1/2 - \eta/4}{1/2 + \eta/4} \right)
\geq \frac{2}{3}  \eta^2,
\end{align*}
where the second to last, and the last inequality follow from $\log x \geq 1 - 1/x$ and $\eta \in [0,1]$, respectively. 
This concludes the proof of inequality~\eqref{eq:dpapleeta}. 
 

\subsection{Proof of Theorem~\ref{thm:parametric}}

The proof is analogous to that of the proof of Theorem~\ref{thm:lowerboundmoderateconf} in Section~\ref{sec:altcmp}, and only requires minor changes.
Specifically, we only need to show that for a given pairwise comparison matrix 
$\Pmat \in \parfamily \cap \comparisonclass[\pmatmin]$, we can construct an alternative matrix obeying equality~\eqref{eq:constraltmtx}, that lies in $\parfamily \cap \comparisonclass[\pmatmin]$ as well. 

Consider any parametric pairwise comparison matrix $\Pmat \in
\parfamily \cap \comparisonclass[\pmatmin]$. 
Then there exists a parameter vector $\vw \in
\reals^\numitems$ such that $\pmat_{ij} = \MYCDF(\parw_i - \parw_j)$.
For the items $\it,\alt \in [\numitems]$, in the proof of Theorem~\ref{thm:lowerboundmoderateconf}, define a set of alternative parameters as
\begin{align*}
\parw_i' & \defeq 
\begin{cases} 
\parw_\alt & \qquad \mbox{if $i=\it$ }, \\ 
\parw_i & \qquad \mbox{otherwise}.
\end{cases}
\end{align*}
Now let $\Pmat'$ be the matrix with pairwise comparison probabilities
$\pmat_{ij}' = \MYCDF(\parw_i' - \parw_j')$. Note that $\pmat' \in \parfamily \cap \comparisonclass[\pmatmin]$, and observe that it obeys equality~\eqref{eq:constraltmtx}, as desired. 

Thus, the proof of Theorem~\ref{thm:lowerboundmoderateconf} yields that for any item $\it$, when applied to a given pairwise comparison model $\pmat \in \parfamily \cap \comparisonclass[\pmatmin]$, the algorithm $\alg$ must make at least 
\begin{align*}
&\frac{2}{3}  \left( \frac{2q -1}{2\h + q} \right)^2
\Big/
\left(
\max_{\alt \in 
\{\setb - 2(\h+q), \ldots, \setb + 1 +  2(\h + q)\} 
}
\max 
\left(
\max_{ j \neq \{\it,\alt\} }
\kl(\pmat_{\it j}, \pmat_{\alt j})
,
\kl(\pmat_{\it \alt}, 1/2)
\right)
\right) \nonumber \\
&\leq 
\frac{2}{3}  \left( \frac{2q -1}{2\h + q} \right)^2
\Big/
\left(
\frac{2\pdfmax^2}{\pmatmin \pdfmin^2} (\score_\it -
\score_\alt)^2
\right) 
\end{align*}
comparisons on average. 
Here, the last inequality follows from \cite[Eq.~(31)]{heckel_active_2016}, which holds for any $i\in [\numitems]$:
\begin{align}
\kl(\pmat_{i \it}, \pmat_{i \alt}') 
%
& \leq \frac{2\pdfmax^2}{\pmatmin \pdfmin^2} (\score_\it -
\score_\alt)^2.
\label{eq:userelwtau} 
\end{align}
Moreover, we used that
\begin{align}
\kl(\pmat_{\it \alt}, 1/2)
& \leq \frac{2\pdfmax^2}{\pmatmin \pdfmin^2} (\score_\it -
\score_\alt)^2,
\label{eq:userelwtau} 
\end{align}
which follows along the lines as~\cite[Eq.~(31)]{heckel_active_2016}. 
This concludes the proof.
 

\section{Discussion}
\label{sec:discussion}

In this paper, we considered the problem of finding an
Hamming-approximate ranking from pairwise comparisons.  We provided an
algorithm that allows to significantly reduce the sample complexity if
one is content with an approximate ranking.  Moreover, we showed that
our algorithm is near optimal and remains near optimal when imposing
common parametric assumptions.  There are a number of open and
practically relevant questions suggested by our work.  As our work
shows, it is non-trivial to adapt to approximate notions of
ranking. It would be interesting to further understand how one can
optimally adapt to approximate notions of ranking, by closing the gap
of our bounds for pathological problem instances, and more importantly
by studying other notions of approximate rankings. It would also be
interesting to study algorithms that work with a limited budget of
queries and quantify their approximation accuracy.


\subsubsection*{Acknowledgements}

The work of RH was supported by the Swiss National Science Foundation
under grant P2EZP2\_159065.  
The work of MJW was partially supported by grants DOD ONR-N00014 and NSF-DMS-1612948.


\printbibliography


\appendix
 
\section{\label{sec:proofequplorel}Proof of equation~\eqref{eq:UpperLowerBoundRel}}

Equation~\eqref{eq:UpperLowerBoundRel} follows by upper bounding the terms in
\[
\Nup{\h}{\pmat}
= 
\Ohidinglogs \left(
\sum_{i = 1}^{\setb-3\h}\gap{i}{\setb+1+3\h}^{-2} 
+
\sum_{i = \setb+1+3\h}^{\numitems} \gap{\setb-3\h}{i}^{-2}
+
2(3\h) \gap{\setb-3\h}{\setb+1+3\h}^{-2}
\right).
\]
Specifically, if $i_1 < i_2$ and $j_2 > j_1$, then $\gap{i_1}{j_1} \le \gap{i_1}{j_1}$. 
Therefore, the terms above can be upper bounded by 
\begin{align*}
\sum_{i = 1}^{\setb-3\h}\gap{i}{\setb+1+3\h}^{-2} \le \sum_{i = 1}^{\setb-3\h}\gap{i}{\setb+1+2\h}^{-2}
,
\quad
\sum_{i = \setb+1+3\h}^{\numitems} \gap{\setb-3\h}{i}^{-2} \le \sum_{i = \setb+1+3\h}^{\numitems} \gap{\setb-2\h}{i}^{-2}, \\
\text{ and } 
2\h \gap{\setb-3\h}{\setb+1+3\h}^{-2} \le \sum_{i=\setb-3\h + 1}^{\setb-2\h} \gap{i}{\setb-2\h + 1}^{-2} + \sum_{i = \setb + 2\h+1}^{\setb - 3\h}\gap{\setb-2\h}{i}^{-2}.
\end{align*}


\end{document}